\documentclass{article}

\usepackage[a4paper, total={6in, 10in}]{geometry}
\usepackage{graphicx} 
\usepackage{gensymb} 
\usepackage{multicol}
\usepackage{tablefootnote} 
\usepackage{booktabs}
\usepackage{amsmath}
\usepackage{gensymb}
\usepackage{xcolor}
\usepackage{comment}
\usepackage{multirow}
\usepackage{hyperref}
\usepackage{makecell}
\usepackage{tabularx}
\usepackage{array}

\usepackage{authblk}

\usepackage{float}

\title{IberFire - a detailed creation of a spatio-temporal dataset for wildfire risk assessment in Spain}

\author[1, 2]{Julen Erzibengoa Calvo}
\author[1]{Meritxell Gómez-Omella}
\author[2]{Izaro Goienetxea Urkizu}

\affil[1]{Tekniker, Spain\\
  \texttt{\{julen.ercibengoa, meritxell.gomez\}@tekniker.es}
}
\affil[2]{EHU\\
  \texttt{jercibengoa001@ikasle.ehu.eus, izaro.goienetxea@ehu.eus}
}

\date{}

\begin{document}

\maketitle

\begin{abstract}

Wildfires pose a threat to ecosystems, economies and public safety, particularly in Mediterranean regions such as Spain. Accurate predictive models require high-resolution spatio-temporal data to capture complex dynamics of environmental and human factors. To address the scarcity of fine-grained wildfire datasets in Spain, we introduce IberFire: a spatio-temporal dataset with 1 km × 1 km × 1-day resolution, covering mainland Spain and the Balearic Islands from December 2007 to December 2024. IberFire integrates 120 features across eight categories: auxiliary data, fire history, geography, topography, meteorology, vegetation indices, human activity and land cover. All features and processing rely on open-access data and tools, with a publicly available codebase ensuring transparency and applicability. IberFire offers enhanced spatial granularity and feature diversity compared to existing European datasets, and provides a reproducible framework. It supports advanced wildfire risk modelling via Machine Learning and Deep Learning, facilitates climate trend analysis, and informs fire prevention and land management strategies. The dataset is freely available on Zenodo to promote open research and collaboration.
\end{abstract}

\section{Background \& Summary}

Forest fires constitute a critical environmental issue with severe ecological, social, and economic implications. Wildfires not only destroy vast forest areas, cause the loss of natural habitats, and release large amounts of carbon dioxide, but also cause substantial economic damage through the destruction of infrastructure, housing, and productive land.

\begin{figure}
    \centering
    \includegraphics[width=0.8\linewidth]{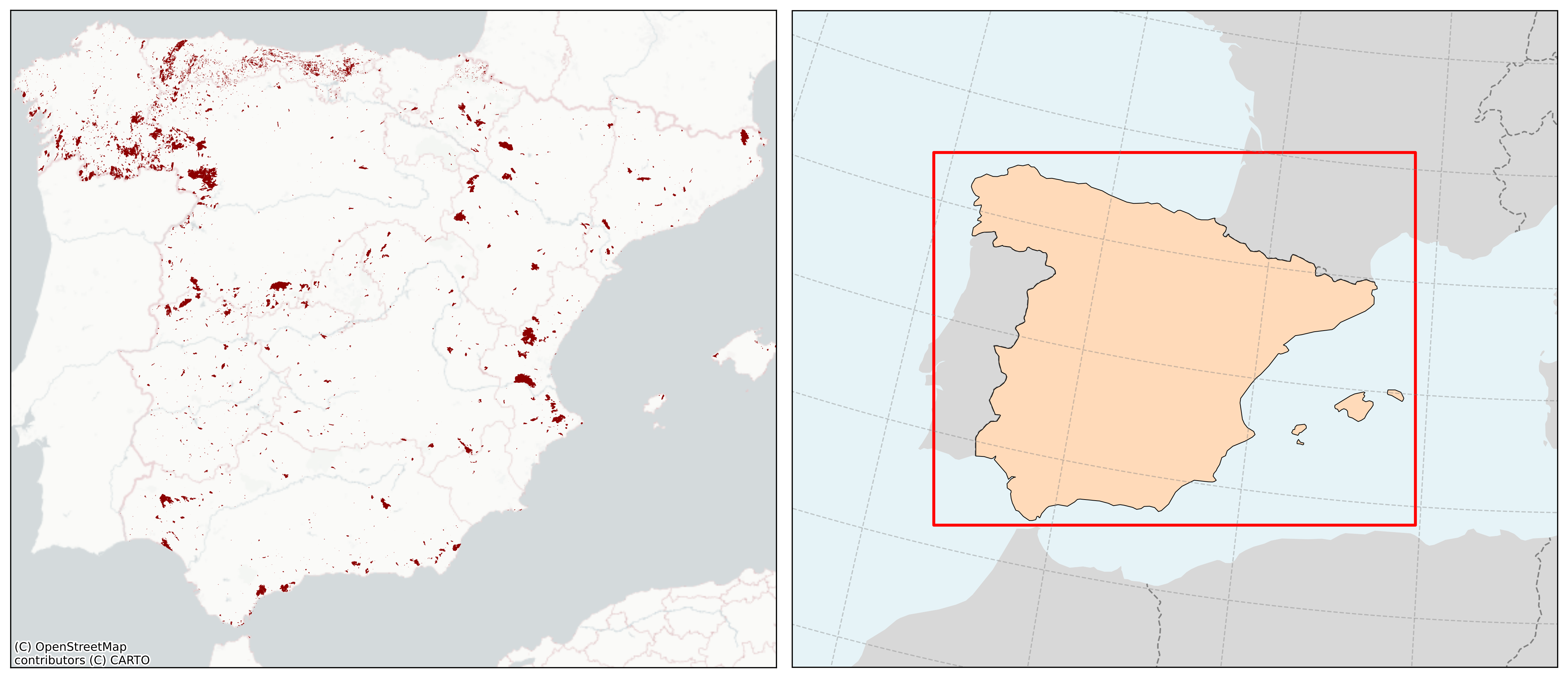}
    \caption{Left: wildfires that occurred from 2008 to 2024, according to data from EFFIS. Right: selected area of interest to build the datacube.}
    \label{fig:fires_and_area_of_interest}
\end{figure}

Spain is one of the countries most affected within the European Union \cite{rodrigues_insight_2014, oliveira_modeling_2012}. Nearly 40\% of the total burned area in the entire Mediterranean region of Europe between 1980 and 2008 was in Spain \cite{moreno2012modelo}. Furthermore, data from the European Forest Fire Information System (EFFIS) \cite{effis} indicate that over 7,000 fires have occurred in Spain since 2008, as shown in the left image of Figure \ref{fig:fires_and_area_of_interest}. 

Wildfires are increasing in scale, with a growing number of autonomous communities experiencing wildfires spanning areas of 5,000 ha, 10,000 ha, and even 20,000 ha \cite{ubeda_grandes_2023}. The year 2022 marked the most severe wildfire season, with two consecutive forest fires jointly burning over 60,000 ha in the same region, an area roughly equivalent to that of Madrid.

In this context, the development of precise predictive models can support fire prevention and fire service management by providing early warnings and identifying high-risk areas. Physical models, such as the Canadian Fire Weather Index \cite{FWI}, leverage meteorological data and physical equations to make predictions; however, recent studies have shown that data-driven models often achieve superior performance in terms of predictive accuracy \cite{prapas_deep_2021,kondylatos_wildfire_2022}.

Building Machine Learning (ML) and Deep Learning (DL) models for fire risk assessment requires high-resolution spatio-temporal features. Many data sources are available for this purpose: the Corine Land Cover (CLC) \cite{CLC} dataset provides a classification of land usage, ERA5-Land \cite{ERA5_validation} offers hourly meteorological curated data and vegetation indices can be retrieved from the Copernicus Land Monitoring Service (CLMS) \cite{CopernicusLandMonitoringService}. When complemented with additional variables, these data enable the development of robust fire-risk prediction models. However, these sources differ significantly in spatial and temporal resolution, format, and update frequency, making direct integration a non-trivial task. 

Datacubes are multidimensional data structures designed to standardise spatial and spatio-temporal features with varying original resolutions, providing a consistent and accessible means of analysis. They facilitate the modelling of complex spatio-temporal phenomena eliminating the need for independent processing pipelines for each data source. In the context of wildfire risk prediction, datacubes are particularly crucial, as they enable the integration of historical fire records alongside heterogeneous environmental variables that influence fire behaviour, like the CLC dataset and ERA5-Land.

To the best of current knowledge, only two datacubes that include Spain within their area of interest are available for this purpose, although neither is specifically focused on the Spanish territory. On the one hand, \textit{SeasFire Cube} \cite{alonso_2024_13834057} offers 59 variables from 2001 to 2021 with 0.25$\degree$ spatial resolution (approximately 27 km at the equator) and 8-day temporal resolution. While this dataset may be used for fire-risk predictions, its resolution is likely too coarse for practical use at the scale of Spain. On the other hand, \textit{Mesogeos} \cite{kondylatos2023mesogeos} offers a 1 km spatial resolution covering the Mediterranean area from 2006 to 2022, with 27 spatio-temporal features. In this case, it is believed that incorporating a broader range of Spain-specific features could enhance forest fire risk predictive models, potentially improving the accuracy of the predictions.

The \textit{IberFire} \cite{erzibengoa_2025_15225886} datacube was constructed to address this gap. It is a 1km $\times$ 1km $\times$ 1-day high-resolution datacube covering Spain from December 2007 to December 2024. It includes 120 features identified in the literature as relevant to forest fire risk. All features were selected based on their potential to be automatically retrieved from external sources, allowing for real-time model deployment.

The features of \textit{IberFire} can be divided into 8 main categories: \textbf{auxiliary features} that assist in locating the cells, \textbf{fire history} obtained from EFFIS, \textbf{geographical location information} features, \textbf{land usage} from Copernicus Corine Land Cover \cite{CLC}, \textbf{topography variables} obtained from the European Digital Elevation Model \cite{EU_DEM}, \textbf{human activity} related features retrieved from WorldPop\cite{worldpop} and OpenStreetMap \cite{openstreetmap}, \textbf{meteorological variables} obtained from ERA5-Land \cite{ERA5_validation}, and \textbf{vegetation indices} downloaded from Copernicus Land Monitoring Service \cite{CopernicusLandMonitoringService}. 

This paper presents two main objectives. The first is the introduction and public release of the \textit{IberFire} datacube, which improves upon existing datasets in terms of resolution and feature diversity for Spain. The \textit{IberFire} datacube offers high-resolution modelling capabilities to gain insights not only into Spain's fire risk behaviour but also into time-series modelling of climate change patterns. The second objective is the provision of a reproducible, systematic methodology for constructing similar datacubes, an approach that can be extended to model other spatio-temporal environmental phenomena. The presented methodology includes a detailed explanation of the generation of \textit{IberFire}, along with the concepts needed to manipulate geospatial data.

\subsection{Concepts and tools about Geographic Information Systems (GIS)} \label{sec_1:concepts_and_tools}

Geographic features are not usually stored in commonly used formats such as CSV (Comma Separated Values); instead, specific formats that contain integrated geographical coordinates, are used; \textit{raster} data is an example of this. Among these formats, \textit{datacubes} organise spatial and temporal information into structured, multi-dimensional arrays. This structure enables efficient storage, retrieval, and analysis of geospatial variables over both space and time, since any cell of the grid can be accessed and every feature value of that cell can be retrieved. The construction of such a datacube requires an understanding of Geographic Information Systems (GIS). This subsection provides an introduction to GIS, emphasising the main data formats and processing techniques involved in manipulating spatial data.

Geographic Information Systems comprise a wide range of tools and data formats specifically designed for the storage, management, and visualisation of spatially-referenced data. In GIS, data are primarily represented using two distinct formats: vector and raster.

\begin{figure}
    \centering
    \includegraphics[width=\linewidth]{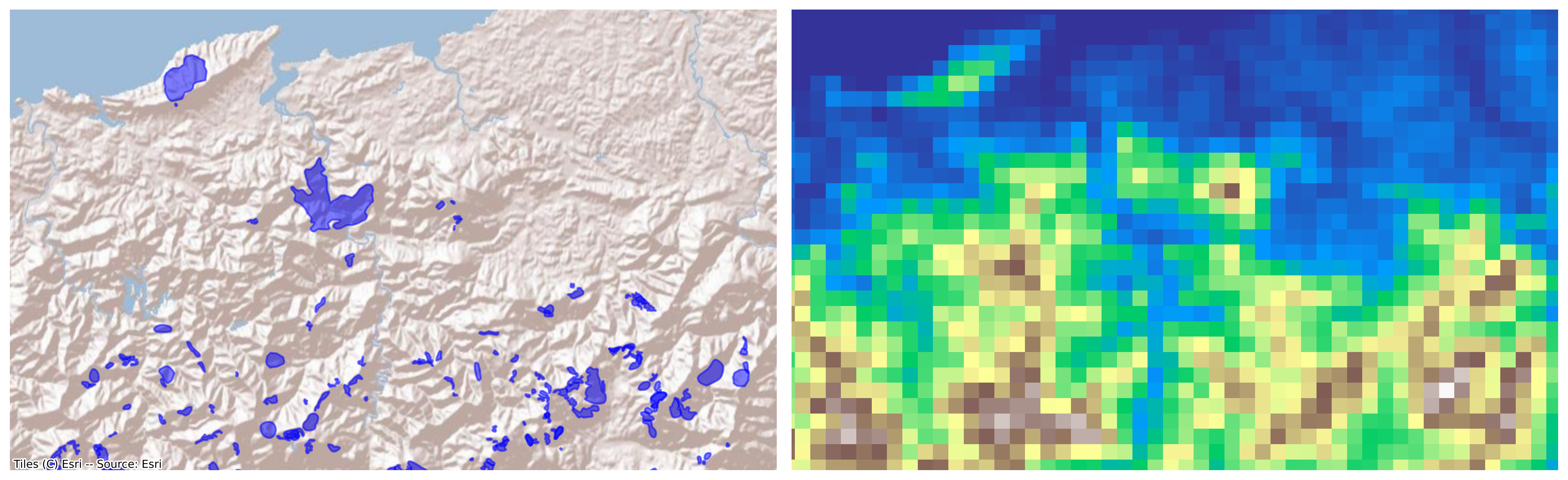}
    \caption{Left: Example of vectorial data (blue), some burned areas retrieved from EFFIS. Right: Example of raster data, elevation values on the same region as the left plot at a 1km$\times$1km resolution.}
    \label{fig:vector_vs_raster}
\end{figure}

Vector data represents geographic features using points, lines, and polygons, which accurately capture geometric locations and boundaries \cite{QGIS_VectorAttributeData}. Each instance of a vector dataset corresponds to a geographic shape with some feature values assigned. This representation is particularly suited for discrete features such as roads or specific fire-affected areas, as exemplified in the left image of Figure \ref{fig:vector_vs_raster}. 

On the other hand, raster data represents geographic space as a regular grid of cells \cite{QGIS_raster_data}, where each pixel is assigned a value corresponding to a property of the geographic area, as can be seen in the right image of Figure \ref{fig:vector_vs_raster}. This data format is commonly used to represent continuous geographic phenomena such as climate data.

Interpolation methods are commonly used to adjust the resolution of raster data when homogenising datasets, a process known as \textit{resampling}. For instance, this process can involve converting data from a finer spatial resolution, such as 100m $\times$ 100m, to a coarser resolution, like 1 km $\times$ 1 km. One widely used interpolation technique is the nearest neighbour interpolation, which assigns the value of the closest input cell to the output cell. Another resampling technique is average resampling, which computes the mean of all input cells that fall within the extent of each output cell.

Geographic data, whether in vector or raster format, relies on Coordinate Reference Systems (CRS). A CRS provides a standardised framework for accurately representing locations on the Earth's surface. Different CRSs are designed to minimise spatial distortion, depending on the specific geographic area and the purpose of the analysis. For instance, the WGS84 CRS (also known as EPSG:4326), which is based on latitude and longitude, is often used for global analysis. In contrast, ETRS89-LAEA (also known as EPSG:3035) can be used for analysis based in Europe since it uses metres as units. 

Downloaded GIS data may come in different CRSs, therefore, it is essential to transform all datasets to a common CRS, a process known as \textit{reprojection}.  In this study, all spatial layers were reprojected to the EPSG:3035 coordinate system, which is particularly suited for European spatial analyses due to its equal-area properties. However, individual CRS transformations applied during the preprocessing stage are not detailed, as they follow standard reprojection procedures widely adopted in geographic data processing.

Effectively processing geospatial data often requires the use of specialised tools. QGIS \cite{QGIS} is an open-source software employed for working with vector and raster data. This software offers an extensive set of functions for visualising, manipulating, automatically reprojecting, and resampling spatial datasets. Other tools for working with GIS data include the \texttt{rasterio} and \texttt{xarray} Python libraries, which provide modules for resampling raster data and creating datacubes, respectively. In this work, both QGIS and Python-based tools were employed, ensuring the reproducibility of the entire process through open-source software solutions.

\subsubsection*{Datacubes}

In GIS, datacubes are essentially raster arrays stacked one on top of the other across time. Hence, for every timestamp, there is one raster representing the values of a specific feature at that timestamp. The rasters share the same CRS, coordinate values and resolution, hence, the only thing that differ are the values of the feature. For every feature a datacube contains, there is a set of stacked raster arrays; and the combination of all the features creates a datacube.

\begin{figure}
    \centering
    \includegraphics[width=0.5\linewidth]{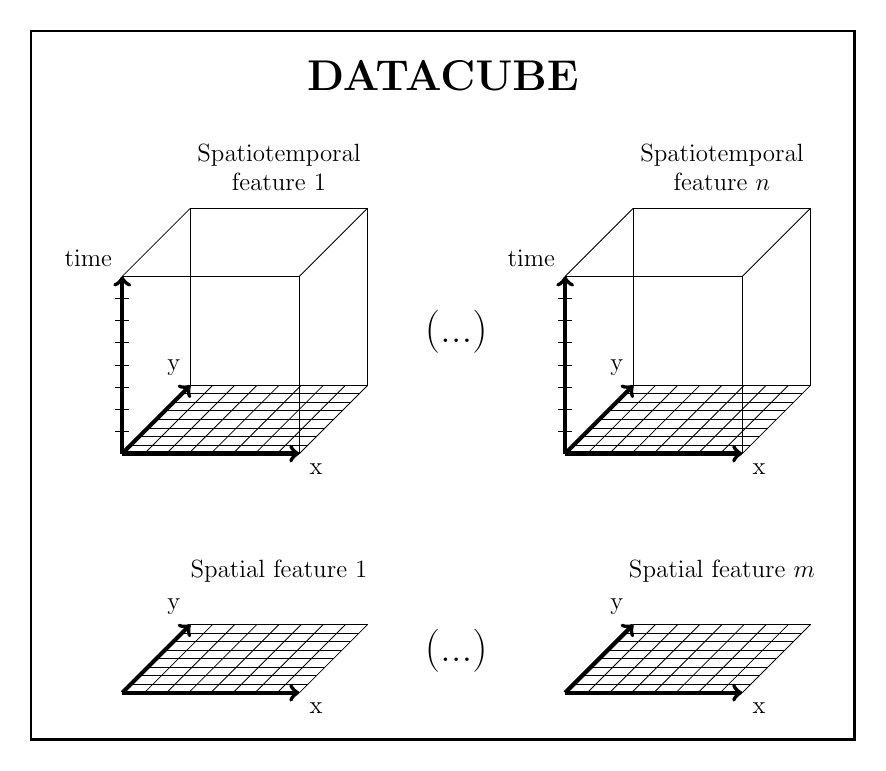}
    \caption{Visual representation of a datacube.}
    \label{fig:datacube_representation}
\end{figure}

As some features do not vary over time, in order not to store repeated values, they can be saved as spatial-only features; hence, only one array is saved for the entire period that the datacube covers. For instance, elevation values could be stored as a spatial-only feature, whereas temperature values should be stored as spatio-temporal. Figure \ref{fig:datacube_representation} provides a visual representation of the datacube concept.

\textit{IberFire} comprises 1188 $x$ coordinates, 920 $y$ coordinates and 6241 timestamps ($t$ or time), one every day. For each $(x,y,t)$ cell, a vector with all the feature values on that specific cell can be retrieved. The values of the retrieved spatial-only features will remain constant if the value of $t$ is changed but $(x,y)$ values do not vary. And the values of the retrieved spatio-temporal features will change if $t$ changes, even if $(x,y)$ remains constant. As an example, elevation does not vary even if $t$ changes, but temperature, stored as a spatio-temporal feature, varies from day to day. 

This storage approach is particularly useful for features with low temporal update frequency. For example, features that update annually—although technically spatio-temporal—can be stored as spatial-only layers. This avoids storing 365 identical rasters per year and instead requires just a single raster array, significantly reducing redundancy and storage requirements.

\section{Methods}\label{chapter:methods}

The creation of a \textit{datacube} involves several steps. First, the spatio-temporal extent and resolution should be defined, which is influenced by the required granularity for the problem, the available computational resources, and the inherent resolution of the data to be employed. Once the spatio-temporal grid has been generated, the desired features should be downloaded and incorporated into the \textit{datacube}. This is the most time-consuming stage, as it is highly feature-specific and requires the careful curation and integration of each variable. To achieve this, data are usually reprojected to a different coordinate reference system, interpolated, and combined.

This section provides a detailed analysis of the construction process of the \textit{IberFire} datacube. Subsection \ref{sec_2:grid_generation} describes the generation of the spatio-temporal grid. After that, Subsection \ref{sec_2:aux_features} analyses the auxiliary features introduced to the datacube. Then, Subsection \ref{sec_2:fire_instances} presents a thorough explanation of the integration of the primary output feature, \texttt{is\_fire}, alongside the integration of the baseline model, the Fire Weather Index (\texttt{FWI}). Subsequently, Subsections \ref{sec_2:geographical_location_information} to \ref{sec_2:vegetation_indices} describe the incorporation of the explanatory variables into the dataset. Finally, Subsection \ref{sec_2:external_data} provides a summary of all external sources utilised in the construction of the datacube.

\subsection{Grid generation} \label{sec_2:grid_generation}
Given that datacubes are many equal-shaped raster arrays arranged along the time dimension, the grid of a datacube is required to be rectangular. A key consideration in the creation of the grid was ensuring that each cell represented exactly 1 km$^2$ of area, which ensures that all cells have the same importance and no cell is underrepresented. To achieve this, the grid was created on the EPSG:3035 CRS, whose units are metres. Within QGIS, the region of interest shown in the right image of Figure \ref{fig:fires_and_area_of_interest} was selected, and an empty base raster file with a 1km $\times$ 1km spatial resolution was generated on that region. This empty raster file was then saved as a reference for the subsequent generation of the datacube and the creation of all the layers it contains. 

The datacube was then constructed in Python using the \texttt{xarray} package. To build a datacube, it is necessary to define coordinate values in each dimension, hence, for $x$, $y$, and $t$ dimensions. Each coordinate consists of a vector of values that are used to locate the individual grid cells. These coordinates are similar to indices, but they can be any ordered set of values, including temporal sequences such as dates. 

The base raster file's horizontal and vertical coordinate values served as the coordinates of the $x$ and $y$ dimensions of the datacube respectively. Then, for the temporal dimension, the coordinates consisted of daily timestamps from 01/12/2007 to 31/12/2024. The final datacube comprised 1188 values for the $x$ dimension, 920 values for the $y$ dimension, and 6241 values for the time axis. As a result, the datacube contains a total of $1188\cdot920\cdot6241 \approx 6.8\cdot 10^9$ different cells, each storing different features. However, not all of these cells should be used for modelling, since not all the cells fall inside Spain, as mentioned in Section \ref{chapter:usage_notes}. The amount of usable cells is $498530 \cdot 6241 \approx 3.1\cdot10^9$.

One important remark is that when adding new features to the datacube, it is possible to select the dimensions that affect the feature. For example, a feature that remains constant over time would only require the $x$ and $y$ dimensions, excluding the time axis, as represented in Figure \ref{fig:datacube_representation}. This approach prevents redundant storage of repeated values for time-invariant features.

\subsection{Auxiliary features} \label{sec_2:aux_features}
After constructing the empty datacube, three auxiliary features were added to the dataset: \texttt{x\_index}, \texttt{y\_index}, and \texttt{is\_spain}. These features, as explained in more detail in Section \ref{chapter:usage_notes}, are not intended to serve as explanatory variables, but rather to facilitate the manipulation and processing of the datacube. 

The first two auxiliary variables, \texttt{x\_index} and \texttt{y\_index}, were introduced to facilitate the identification of grid cells. The \texttt{x\_index} variable consists of a sequence of integer values representing the horizontal index of each cell, ordered from left to right. Similarly, the \texttt{y\_index} indicates the vertical index, ordered from top to bottom. These auxiliary variables are particularly useful when individual cell values are extracted from the datacube and stored in a standard CSV format, as they allow each instance to be accurately mapped back to its corresponding original spatial location within the datacube.

The third auxiliary feature, \texttt{is\_spain}, identifies the cells corresponding to Spanish territory, which are the only ones for which predictions should be generated. This layer was derived from a vectorial dataset containing the boundaries of Spain obtained from simplemaps \cite{simplemaps-spain} and processed with the QGIS software.

Given that these auxiliary features do not vary over time, they were stored as {spatial} features, with only the $x$ and $y$ coordinate values.

\subsection{Fire history: EFFIS} \label{sec_2:fire_instances}
\begin{figure}[!ht]
    \centering
    \includegraphics[width=\linewidth]{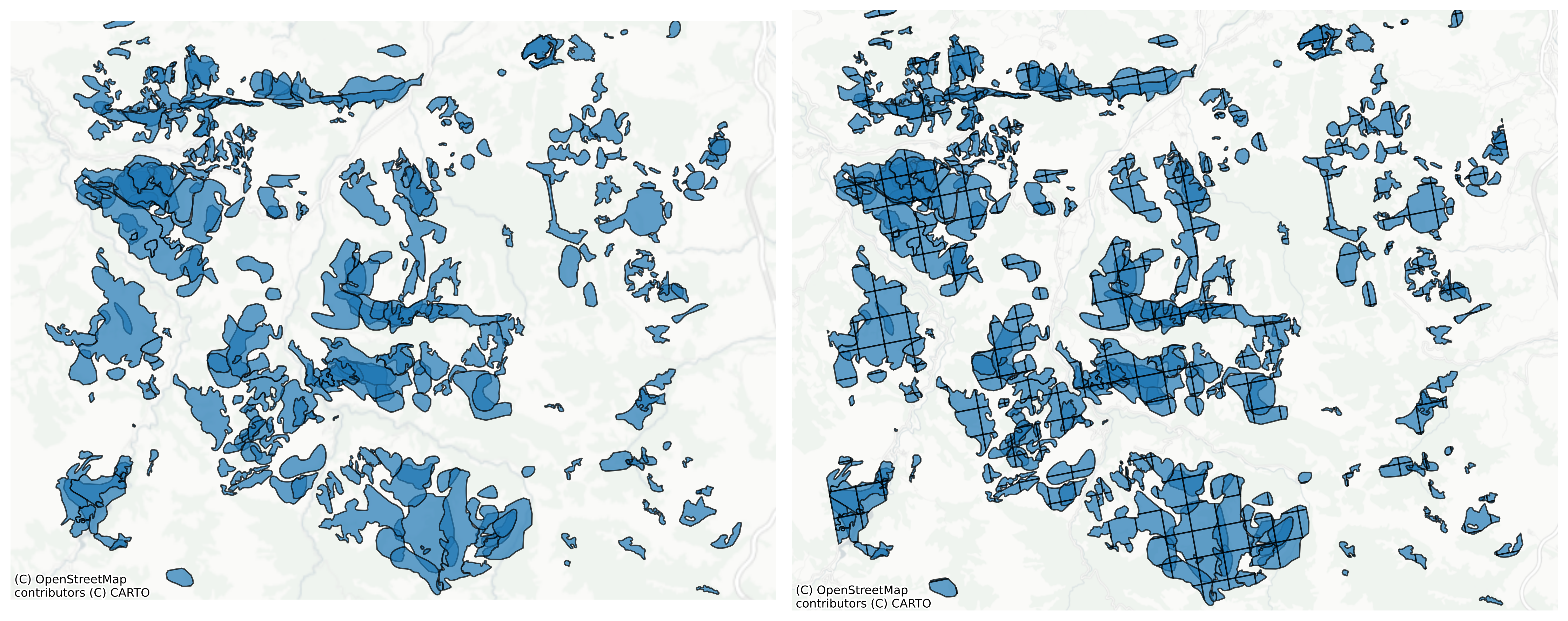}
    \caption{Left: raw geometries of the fire events downloaded from EFFIS. Right: the same geometries intersected with the spatial grid.}
    \label{fig:fires_plot}
\end{figure}

The historical fire data for Spain were obtained using the EFFIS data request format \cite{effis}. The raw data were retrieved in vectorial format and contained geometries representing the burned area of historical fire events, along with the corresponding start and end dates of each fire event. To integrate this information into the datacube, the intersection of the spatial grid cells with the fire geometries in QGIS was calculated, as illustrated in Figure \ref{fig:fires_plot}. Subsequently, a binary layer in the datacube, named \texttt{is\_fire}, was created. A value of 1 was assigned to cells that intersected with fire geometries and fell within the corresponding temporal interval defined by the fire's start and end dates. Following the definition of fire danger from \cite{prapas_deep_2021}, the previous process was executed on fires (geometries retrieved from EFFIS) with a burned area greater than 5 ha. This is because fire danger can be viewed as the combined risk of a fire igniting and the risk of that fire growing large (\textgreater 5 ha). Introducing small wildfires as fire instances (\texttt{is\_fire} = 1) could lead to inconsistencies, as these small fires did not grow larger for certain reasons (for example, high humidity). Therefore, the fire risk for these small fires was low, even though a fire was present.

After that, the \texttt{is\_near\_fire} layer was added, which is a binary feature indicating whether a cell is within a 25 $\times$ 25 cell area centred on each \texttt{is\_fire} = 1 instance, as well as the 10 days preceding the event. This results in a 25 $\times$ 25 $\times$ 10-sized box prior to each fire cell. The \texttt{is\_near\_fire} feature is particularly useful for identifying true non-fire events that are neither spatially nor temporally close to fire events, thereby avoiding the inclusion of near-fire instances that may closely resemble actual fire conditions due to their proximity.

Both of these features were added to the datacube as {spatio-temporal} features, hence, with all three $x$, $y$, and $t$ dimensions. 

\subsubsection*{Baseline model: Fire Weather Index} \label{sec_2:fwi_baseline_model}
The \textit{Fire Weather Index} (\texttt{FWI}) \cite{FWI} was introduced to \textit{IberFire} as a baseline model. It leverages temperature, wind, relative humidity and precipitation data, along with physical equations to predict a continuous value that represents fire risk. The \texttt{FWI} takes values in the range $[0, +\inf)$, although the most common values lie between 0 and 50. Depending on the value that it has, fire risk levels are assigned according to Table \ref{tab:FWI_risk_levels} \cite{FWI_explanations}.

\begin{table}[ht]
\centering
\setlength{\tabcolsep}{4pt} 
\renewcommand{\arraystretch}{1.0} 
\begin{tabular}{@{}cccccccc@{}}
\toprule
\textbf{Very low} & \textbf{Low} & \textbf{Moderate} & \textbf{Hight} &
\textbf{Very high} & \textbf{Extreme} \\
\midrule
$<$ 5.2 & 5.2 - 11.2 & 11.2 - 21.3 & 21.3 -38.0 &
38.0 - 50 & $\geq$ 50  \\
\bottomrule
\end{tabular}
\caption{FWI risk levels.}
\label{tab:FWI_risk_levels}
\end{table}

To introduce \texttt{FWI} to \textit{IberFire}, data from the Copernicus Emergency Management Service (CEMS) \cite{FWI_data} was downloaded, which provides daily values at a spatial resolution of 0.25º × 0.25º latitude-longitude, equivalent to approximately 27.5 km. Since the original dataset resolution is lower than desired, the data was interpolated to a 1 km × 1 km resolution using nearest-neighbor interpolation. This feature was also added as a spatio-temporal feature.

\subsection{Geographical location information}\label{sec_2:geographical_location_information}
The geographical location information included in the \textit{IberFire} datacube plays a crucial role. As illustrated in Figure \ref{fig:fires_and_area_of_interest}, certain regions inherently present a higher susceptibility to wildfires than others. Therefore, the inclusion of features representing the spatial position of each cell was deemed advantageous for the models.

Five features were added for this purpose: \texttt{x\_coordinate}, \texttt{y\_coordinate}, \texttt{is\_sea}, \texttt{is\_waterbody} and \texttt{AutonomousCommunities}. Since these geographical location features are time-invariant, they were stored in the datacube using only the spatial dimensions.

The first two features consist of the values of the coordinates of each cell in the EPSG:3035 coordinate reference system. The next two features, \texttt{is\_sea} and \texttt{is\_waterbody}, are binary indicators that denote whether a given cell is located over open sea or inland water, respectively. These features were calculated using QGIS with data from the European Digital Elevation Model \cite{EU_DEM}. 

Lastly, the \texttt{AutonomousCommunities} feature represents the level 2 NUTS (Nomenclature of Territorial Units for Statistics) \cite{nuts_ec_2016} division of Spain, with values listed in Table \ref{tab:community_codes}. However, since the region of interest does not include Ceuta, Melilla, and the Canary Islands, the corresponding values for these regions do not appear in the dataset. The feature was generated in QGIS by converting a vector dataset containing the shapes of the autonomous communities of Spain \cite{area2018limites} into a raster layer, assigning to each cell the corresponding value from Table \ref{tab:community_codes} based on its intersection with the appropriate autonomous community.

\begin{table}[!h]
    \centering
    \begin{tabular}{|ll|ll|}
        \hline
        \textbf{Value} & \textbf{Region} & \textbf{Value} & \textbf{Region} \\
        \hline
        0 & Nodata                & 10 & Comunidad Valenciana \\
        1 & Andalucía             & 11 & Extremadura           \\
        2 & Aragón                & 12 & Galicia               \\
        3 & Principado de Asturias & 13 & Comunidad de Madrid   \\
        4 & Islas Baleares        & 14 & Región de Murcia      \\
        5 & Canarias              & 15 & Comunidad Foral de Navarra \\
        6 & Cantabria             & 16 & País Vasco            \\
        7 & Castilla y León       & 17 & La Rioja              \\
        8 & Castilla - La Mancha  & 18 & Ceuta                 \\
        9 & Cataluña              & 19 & Melilla               \\
        \hline
    \end{tabular}
    \caption{Values of the feature \texttt{AutonomousCommunities} and their corresponding regions.}
    \label{tab:community_codes}
\end{table}

\subsection{Land usage: Corine Land Cover}\label{sec:CorineLandCover_explanations}
The Copernicus Corine Land Cover (CLC) dataset \cite{CLC} offers a standardised classification of land cover types across Europe, distinguishing 44 discrete categories. It represents the continent as a regular grid of 100m $\times$ 100m cells, assigning to each cell an integer value between 1 and 44 that corresponds to a specific land cover class. 

As detailed in Table \ref{table:clc_legend}, each of the 44 land cover classes is associated with three hierarchical labels that facilitate their aggregation into broader thematic groups. The third label, {Label 3}, is the most specific, assigning a unique identifier to each of the 44 classes, therefore comprising 44 distinct categories. {Label 2} serves as an intermediate level, grouping related {Label 3} classes into broader categories, while {Label 1} represents the highest level of aggregation, clustering the 44 land classes into 5 major land cover types.

\begin{table}[!h]
\centering
\tiny
\begin{tabular}{p{0.3cm} p{3.1cm} p{5.1cm} p{4.5cm}}
\hline
\textbf{Class} & \textbf{Label 1} & \textbf{Label 2} & \textbf{Label 3} \\
\hline
1 & Artificial surfaces & Urban fabric & Continuous urban fabric \\
2 & Artificial surfaces & Urban fabric & Discontinuous urban fabric \\
3 & Artificial surfaces & Industrial, commercial and transport units & Industrial or commercial units \\
4 & Artificial surfaces & Industrial, commercial and transport units & Road and rail networks and associated land \\
5 & Artificial surfaces & Industrial, commercial and transport units & Port areas \\
6 & Artificial surfaces & Industrial, commercial and transport units & Airports \\
7 & Artificial surfaces & Mine, dump and construction sites & Mineral extraction sites \\
8 & Artificial surfaces & Mine, dump and construction sites & Dump sites \\
9 & Artificial surfaces & Mine, dump and construction sites & Construction sites \\
10 & Artificial surfaces & Artificial, non-agricultural vegetated areas & Green urban areas \\
11 & Artificial surfaces & Artificial, non-agricultural vegetated areas & Sport and leisure facilities \\
12 & Agricultural areas & Arable land & Non-irrigated arable land \\
13 & Agricultural areas & Arable land & Permanently irrigated land \\
14 & Agricultural areas & Arable land & Rice fields \\
15 & Agricultural areas & Permanent crops & Vineyards \\
16 & Agricultural areas & Permanent crops & Fruit trees and berry plantations \\
17 & Agricultural areas & Permanent crops & Olive groves \\
18 & Agricultural areas & Pastures & Pastures \\
19 & Agricultural areas & Heterogeneous agricultural areas & Permanent crops\tablefootnote{Complete description shortened.}  \\
20 & Agricultural areas & Heterogeneous agricultural areas & Complex cultivation patterns \\
21 & Agricultural areas & Heterogeneous agricultural areas & Land principally occupied by agriculture \color{blue}\footnotemark[1]\color{black}\\
22 & Agricultural areas & Heterogeneous agricultural areas & Agro-forestry areas \\
23 & Forest and semi natural areas & Forests & Broad-leaved forest \\
24 & Forest and semi natural areas & Forests & Coniferous forest \\
25 & Forest and semi natural areas & Forests & Mixed forest \\
26 & Forest and semi natural areas & Scrub and/or herbaceous vegetation associations & Natural grasslands \\
27 & Forest and semi natural areas & Scrub and/or herbaceous vegetation associations & Moors and heathland \\
28 & Forest and semi natural areas & Scrub and/or herbaceous vegetation associations & Sclerophyllous vegetation \\
29 & Forest and semi natural areas & Scrub and/or herbaceous vegetation associations & Transitional woodland-shrub \\
30 & Forest and semi natural areas & Open spaces with little or no vegetation & Beaches, dunes, sands \\
31 & Forest and semi natural areas & Open spaces with little or no vegetation & Bare rocks \\
32 & Forest and semi natural areas & Open spaces with little or no vegetation & Sparsely vegetated areas \\
33 & Forest and semi natural areas & Open spaces with little or no vegetation & Burnt areas \\
34 & Forest and semi natural areas & Open spaces with little or no vegetation & Glaciers and perpetual snow \\
35 & Wetlands & Inland wetlands & Inland marshes \\
36 & Wetlands & Inland wetlands & Peat bogs \\
37 & Wetlands & Maritime wetlands & Salt marshes \\
38 & Wetlands & Maritime wetlands & Salines \\
39 & Wetlands & Maritime wetlands & Intertidal flats \\
40 & Water bodies & Inland waters & Water courses \\
41 & Water bodies & Inland waters & Water bodies \\
42 & Water bodies & Marine waters & Coastal lagoons \\
43 & Water bodies & Marine waters & Estuaries \\
44 & Water bodies & Marine waters & Sea and ocean \\
\hline
\end{tabular}
\caption{Definitions of the 44 classes from Corine Land Cover.}
\label{table:clc_legend}
\end{table}

For instance, the class labelled as \textit{Continuous urban fabric} in {Label 3} is grouped under the category \textit{Urban fabric} in {Label 2}, which, in turn, falls under the broader category \textit{Artificial surfaces} in {Label 1}. More specifically, the category \textit{Urban fabric} includes classes 1 and 2, whereas \textit{Artificial surfaces} includes classes 1 through 11. This hierarchical structure enables both detailed analysis and higher-level generalization.

The original spatial resolution of the CLC dataset is 100m $\times$ 100m; consequently, resampling of the data was needed to match the 1km $\times$ 1km resolution of the \textit{IberFire} datacube. Using QGIS, the proportion of each of the 44 classes within each 1km $\times$ 1km cell was calculated. This resulted in 44 features, denoted as \texttt{CLC\_i} for $i = 1, 2, \dots, 44$, with values ranging between 0 and 1.

Given that these features represent proportions and therefore sum up to 1 for each 1km $\times$ 1km cell, and also considering the hierarchical structure of the CLC classification, five additional features were derived corresponding to the higher-level groupings defined in {Label 1}. Specifically, for each higher-level category, its proportion within a cell was computed by summing the relevant \texttt{CLC\_i} variables that fall inside the category. For example, the proportion of \textit{Artificial surfaces} was obtained by aggregating the values of \texttt{CLC\_1} through \texttt{CLC\_11}.

The same procedure was applied to the intermediate level of the hierarchy, {Label 2}, resulting in the creation of 14 additional aggregated features. Although 15 categories exist at this level, the \textit{Pastures} category (corresponding to class 18 in {Label 3}) was excluded, as it consists of a single class and thus provides no added abstraction over the original variable.

To recapitulate, a total of 63 explanatory variables were derived from the CLC dataset and incorporated into \textit{IberFire}: 44 corresponding to the most detailed classification level {Label 3}, 14 to the intermediate level {Label 2}, and 5 to the highest level of aggregation {Label 1}. 

The CLC dataset is updated every six years; accordingly, the 2006 \cite{CLC_2006_version}, 2012 \cite{CLC_2012_version}, and 2018 \cite{CLC_2018_version} editions were used in this work, as the 2024 version was not yet available at the time of writing. As previously discussed, time-invariant features in the datacube can be stored using only spatial dimensions. Given the low update frequency of the CLC dataset, including a temporal dimension would result in unnecessary duplication of values. Therefore, all CLC-derived features were stored without the time axis. To differentiate between editions, the corresponding year was appended to each feature name, as better described in Table \ref{tab:clc2006} of Section \ref{section:data_records}. Each of the three CLC editions contributed 63 variables, this approach ultimately yielded $63 \cdot 3 = 189$ distinct CLC-derived spatial-only features in the datacube. However, since only one CLC edition is relevant for any given $(x,y,t)$ cell, these 189 variables effectively represent just 63 unique features. The selection of the appropriate CLC edition for each cell is described in detail in Section~\ref{chapter:usage_notes}.

\subsection{Topography variables: European Digital Elevation Model}
Topography is a well-established factor influencing the spread and intensity of forest fires. To capture these effects, topographic features were downloaded from the European Digital Elevation Model (EU-DEM) provided by OpenTopography \cite{EU_DEM}. The original EU-DEM dataset offers {elevation} values at a 30m$\times$30m spatial resolution. Using this elevation data, additional variables such as {slope}, {aspect} and {roughness} of the terrain can be derived, all of which are also offered by OpenTopography.

However, the specific methods used by OpenTopography to calculate the {slope} and {roughness} are not documented, leaving the units of these variables undefined. In contrast, the {aspect} is expressed in degrees, ranging from 0\degree to 360\degree, indicating the direction in which the slope of each 30m$\times$30m terrain cell faces, with 0° corresponding to the north and the values increasing clockwise.

The {elevation}, {slope}, {roughness} and {aspect} were downloaded and, for the first three features, the mean and standard deviation in each 1km$\times$1km cell were calculated. Therefore, a total of 6 different features were obtained from {elevation}, {slope}, and {roughness}.  For {aspect}, the feature was discretised into 8 different classes defined in Table \ref{tab:orientacion}. For each of the 1km$\times$1km cell, the proportion of each class inside the cell was calculated, resulting in 8 new features called \texttt{aspect\_i} for $i =1,\dots,8$. Additionally, an extra feature, \texttt{aspect\_NODATA}, was included to represent the proportion of pixels within each 1km$\times$1km cell for which no aspect value was available, which is highly correlated with \texttt{is\_waterbody}. This lack of data typically occurs in areas covered by lakes and rivers, where aspect values are undefined due to the absence of terrain elevation gradients.

\begin{table}[!h]
\centering
\setlength{\tabcolsep}{4pt} 
\renewcommand{\arraystretch}{1.0} 
\begin{tabular}{@{}cccccccc@{}}
\toprule
\textbf{Class 1} & \textbf{Class 2} & \textbf{Class 3} & \textbf{Class 4} &
\textbf{Class 5} & \textbf{Class 6} & \textbf{Class 7} & \textbf{Class 8} \\
\midrule
$(0,\,45]$ & $(45,\,90]$ & $(90,\,135]$ & $(135,\,180]$ &
$(180,\,225]$ & $(225,\,270]$ & $(270,\,315]$ & $(315,\,360]$ \\
\bottomrule
\end{tabular}
\caption{Orientation classes grouped into 45\degree intervals.}
\label{tab:orientacion}
\end{table}

These processes were performed using QGIS software, and the resulting 15 features were subsequently integrated into the datacube, utilizing only spatial coordinates.

\subsection{Human activity}
It is widely recognised that most wildfires are directly related to human activities, either intentionally or accidentally \cite{liz-lopez_spain_2023}. To model this phenomenon, six human-related explanatory variables were considered: {distance to roads}, {distance to waterways}, {distance to railways}, designation within the {Natura 2000 protected network} \cite{miteco_rednatura2000}, {population density}, and {holiday periods}. These variables were used to derive a total of 20 features, which were subsequently integrated into the datacube.

The first two variables, {distance to roads} \cite{distance_to_roads} and {distance to waterways} \cite{distance_to_waterways}, were obtained from WorldPop \cite{worldpop}, which provides data at an approximate spatial resolution of 100m$\times$100m, represented in kilometres. To upscale the data to the target resolution, the mean and standard deviation within each 1km$\times$1km cell were computed using QGIS. These aggregated statistics were then incorporated into the datacube, resulting in four derived features from the original two variables. These features were considered invariant over time and therefore added with only spatial coordinates.

For {distance to railways}, railway vectorial data in Spain were retrieved from OpenStreetMap \cite{openstreetmap}. Using QGIS, a 100m$\times$100m raster layer was generated to represent the distance to the nearest railway geometry from each 100m$\times$100m cell. Then, the same procedure applied to the previous two variables was then applied here as well, and the average and standard deviation were calculated. These two features were then added to the datacube as spatial data.

The Natura 2000 network, a European ecological network for biodiversity conservation, was included due to its relevance to fire data reported by EFFIS. In the vector fire data provided by EFFIS, the proportion of each forest fire that occurred within Natura 2000 protected areas is specified. Using vectorial boundaries of the Natura 2000 network, a binary raster layer was created with QGIS indicating whether each 1km$\times$1km cell falls within the protected area. Since this variable remains constant over time, it was integrated as a spatial layer into the datacube.

{Population density} was incorporated as a proxy for human presence and potential ignition sources. Annual data were obtained from WorldPop for the years 2008 to 2020 \cite{popdens}, due to the unavailability of more recent data. The original data were retrieved with a resolution of approximately 1km $\times$ 1km, and average resampling was applied to match the coordinate values of \textit{IberFire}. Given the annual update frequency of the data, {population density} were stored as spatial features to prevent unnecessary duplication of values, mirroring the process conducted with the CLC-derived features. Consequently, this variable appears as 13 spatial layers in the dataset, but functionally represents a single feature, as only the population density for the relevant year should be selected for each $(x,y,t)$ cell.

Finally, a binary feature representing {holiday periods} was included. This variable accounts for the increased presence of people in rural and natural areas during weekends and public holidays, potentially raising the risk of fire ignition. A cell was flagged as a holiday if the corresponding date was a Saturday, Sunday, or a national or regional public holiday in Spain. To determine public holidays, the Python library \texttt{holidays} \cite{arkadii_yakovets_2025_15169945} was utilised. This library provides holiday dates for various countries and their subdivisions, allowing for precise identification of holidays at the autonomous community level in Spain. Consequently, the \texttt{AutonomousCommunities} feature was employed to assign the appropriate regional holidays to each cell. The resulting \texttt{is\_holiday} binary layer was added as a spatio-temporal feature within the datacube.

\subsection{Meteorological variables: ERA5-Land}
Meteorological conditions are among the most influential factors driving both the ignition and propagation of forest fires. Variables such as {temperature}, precipitation, and wind speed significantly affect fire behaviour and overall risk levels. Therefore, incorporating meteorological data into the datacube is essential for generating robust predictive models.

To account for these dynamics, data from ERA5-Land \cite{ERA5_validation} were integrated, a high-resolution global reanalysis dataset provided by the Copernicus Climate Data Store. ERA5-Land offers hourly meteorological variables at a spatial resolution of 9 km $\times$ 9 km, from 1950 to the present. These variables are obtained as a combination of meteorological measurements and numerical weather prediction models through data assimilation techniques. This approach ensures spatial and temporal consistency.

ERA5-Land is particularly well-suited for environmental modelling due to its global coverage, temporal consistency, and availability of multiple curated atmospheric variables relevant to fire risk assessment. Within the Copernicus Climate Data Store, two data access modes are available: raw hourly observations \cite{era5_hourly_data} and post-processed daily statistics \cite{era5_postprocessed_data}. Both sources were used in the construction of the \textit{IberFire} datacube, depending on the specific requirements of each variable. 

Although ERA5-Land provides temporally consistent data, its availability is subject to a delay of five days, making it unsuitable for real-time fire risk prediction, where daily forecasts are required. Consequently, while ERA5-Land data were used to construct the datacube, real-time model deployment is intended to rely on meteorological station measurements. 

The use of meteorological station measurements for the construction of the datacube was also considered, but the lack of historical data for many variables in various meteorological stations posed significant limitations.

AEMET (the Spanish Meteorological Agency) provides open-access, near real-time data from weather stations across Spain. Therefore, to ensure compatibility, all meteorological ERA5-Land features included in \textit{IberFire} were selected and processed to align with the type and format of data provided by AEMET. 

A total of 17 meteorological features were derived from ERA5-Land data. The features can be grouped as follows: {temperature}, {relative humidity}, {surface pressure}, {precipitations}, {wind speed}, and {wind direction}. This subsection outlines the methodology used to obtain and incorporate these variables into the \textit{IberFire} datacube, including the selection of source variables and the transformations necessary to ensure compatibility with the format and units used in AEMET observations.

ERA5-Land data are available at a global scale; therefore, all relevant variables were extracted for the specific region of interest illustrated in Figure \ref{fig:fires_and_area_of_interest}. To ensure spatial consistency across the entire datacube, all ERA5-Land data, originally provided at a resolution of 9km $\times$ 9km, were resampled to the target 1 km $\times$ 1 km resolution using nearest-neighbour interpolation. All the meteorological features were added to the datacube as spatio-temporal variables, and the processing of these features was done using Python.

\subsubsection*{Temperature}
Temperature plays a central role in wildfire risk modelling, as it directly affects the dryness and flammability of vegetation. Four daily statistics were saved to describe the {temperature}: {mean}, {minimum}, {maximum}, and {range}. The first three features were extracted from the daily statistics of the variable {2m temperature} provided by ERA5-Land, which measures the temperature of the air at 2 metres above the surface of land. Then, the {range}, which is the difference between the daily maximum and minimum value, was calculated. Finally, the original units in Kelvin provided by ERA5-Land were transformed into Celsius, ensuring consistency with the units used by AEMET.

\subsubsection*{Relative humidity}

Relative humidity is the percentage of moisture in the air relative to the maximum amount the air can hold at a given temperature. It is a critical variable for fire risk assessment, as it directly affects the moisture content of vegetation and, consequently, the likelihood of ignition and propagation. However, ERA5-Land does not provide this variable directly, while AEMET does. To address this, hourly relative humidity values were derived from two available ERA5-Land variables: the {2m temperature} and the {2m dewpoint} temperature. The last one represents the temperature at which the air becomes saturated with moisture, at 2 metres above the ground. Using these two temperature variables, relative humidity values were computed by applying the Magnus formula \cite{ImprovedMagnusFormApproximationofSaturationVaporPressure}, an empirically validated approach for estimating saturation vapor pressure:

\begin{equation}\label{formula:RelativeHumidity}
    \textit{Relative Humidity} = \frac{\exp(\frac{17.625 \cdot D_p}{243.04 + D_p})}{\exp(\frac{17.625 \cdot T}{243.04 + T})}
\end{equation}
where $D_p$ and $T$ represent the {2m dewpoint temperature} and {2m temperature} in Celsius, respectively.

From the computed hourly values, four daily statistics were derived and included in the datacube: the {mean}, {minimum}, {maximum}, and {range} of {relative humidity}.

Computing daily relative humidity statistics first required deriving hourly relative humidity values from the corresponding hourly temperature and dewpoint temperature data. This step was essential because daily statistics of relative humidity cannot be accurately obtained from the summary statistics of the temperature variables provided by ERA5-Land. For instance, inserting the daily mean values of $D_p$ and $T$ into Equation~\eqref{formula:RelativeHumidity} does not yield the correct daily mean of relative humidity.

\subsubsection*{Surface pressure}
{Surface pressure} values were retrieved from ERA5-Land as daily aggregated statistics, namely the {mean}, {minimum}, and {maximum}, originally expressed in pascals. Again, the daily {range} was then computed as the difference between the maximum and minimum values. Finally, all four features were converted to hectopascals to ensure consistency with the unit conventions used by AEMET.

\subsubsection*{Precipitations}
Since pre-aggregated {precipitation} statistics were not available, hourly precipitation values were downloaded from ERA5-Land. These values, originally expressed in metres (equivalent to 1000 l/m$^2$), were averaged to obtain daily {mean precipitation} values. Subsequently, the units were converted from metres to millimetres to ensure consistency with AEMET.

\subsubsection*{Wind speed}
Wind-related features required particularly careful processing. ERA5-Land provides wind data in terms of its eastward ({u-wind}) and northward ({v-wind}) components, therefore, two hourly features, u-wind and v-wind, can be retrieved from ERA5-Land. In contrast, AEMET reports wind information as maximum and average wind speeds, without disaggregating it into horizontal and vertical components.

To align the ERA5-Land data with the format used by AEMET, hourly {u-wind} and {v-wind} values were retrieved from ERA5-Land. Then, the wind speed magnitude at each hour was computed using the Euclidean norm, taking into account the orthogonality of the components:
\begin{equation}\label{wind_speed_formula}
\lVert \vec{(u,v)} \rVert = \sqrt{\lvert {u} \rvert^2 + \lvert {v}\rvert^2} \,.
\end{equation}

From these hourly magnitudes, the daily maximum and average wind speeds were calculated and incorporated into the datacube as spatio-temporal features, named \texttt{wind\_speed\_max} and \newline \texttt{wind\_speed\_mean}. The data were stored with the original ERA5-Land units, metres per second (m/s), since they match with the units used by AEMET.

It is important to note that, analogous to the case of {relative humidity}, daily wind speed statistics cannot be accurately computed from the daily maximum or average values of {u-wind} and {v-wind} individually. The magnitude must be calculated at the hourly level before aggregation.

\subsubsection*{Wind direction}

Wind direction is conventionally defined as the direction from which the wind originates, expressed in degrees measured clockwise from true north. AEMET provides two daily wind-direction metrics: the average wind direction and the direction at the daily maximum wind speed. Meanwhile, ERA5-Land offers hourly wind data at a resolution of 9km $\times$ 9km, represented by horizontal ({u\_wind}) and vertical ({v\_wind}) components. To convert these components into standard wind direction, equation \eqref{eq:wind_dir_formula} is employed, where $\alpha$ denotes the angle between $\vec{(u,v)}$ and the north vector $\vec{(0,1)}$.

\begin{equation}\label{eq:wind_dir_formula}
\textit{Wind direction} = (\alpha - 180^\circ)\bmod 360^\circ \,,
\quad
\alpha = \angle\bigl(\vec{(0,1)},\vec{(u,v)}\bigr)\,.  
\end{equation}

\begin{figure}
    \centering
    \includegraphics[width=.75\linewidth]{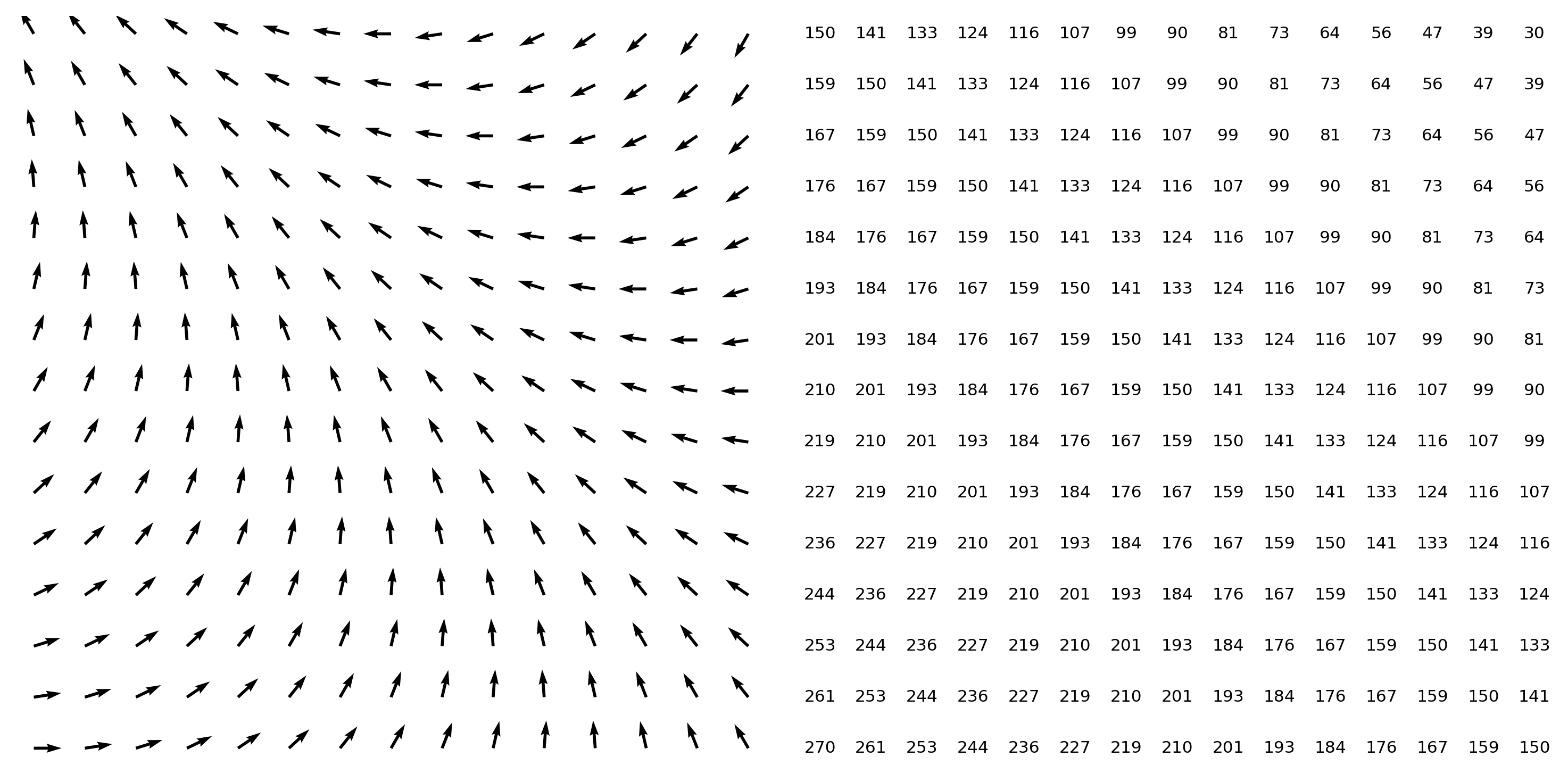}
    \caption{Left: wind vectors that represent the sum of ERA5-Land {u-wind} and {v-wind} components. Right: the wind direction values of those same vectors.}
    \label{fig:wind_direction_values}
\end{figure}

Figure \ref{fig:wind_direction_values} demonstrates this process by comparing the original component-wise data from ERA5-Land on the left, with the derived wind direction map on the right. 

To obtain the daily average wind direction, daily averaged values of {u\_wind} and {v\_wind} were first obtained from ERA5-Land's aggregated statistics and converted with equation \ref{eq:wind_dir_formula}. Subsequently, to capture the direction at each day's maximum wind speed, hourly {u\_wind} and {v\_wind} components were downloaded, and equation \eqref{eq:wind_dir_formula} was applied at each hourly timestamp. The direction corresponding to the highest wind speed of the day was then retained. This resulted in the addition of two spatio-temporal features, \texttt{wind\_direction\_mean} and \texttt{wind\_direction\_at\_max\_speed}.

\subsection{Vegetation indices: Copernicus Land Monitoring Service} \label{sec_2:vegetation_indices}

The last group of features integrated into the datacube is the vegetation indices, which are highly used for making fire-risk predictions, as they can represent the dryness of the plant life. Five different indices were added: \textit{FAPAR} (Fraction of Photosynthetically Active Radiation), \textit{LAI} (Leaf Area Index), NDVI (Normalised Difference Water Index), LST (Land Surface Index), and SWI (Soil Water Index). All of them were retrieved from the Copernicus Land Monitoring Service (CLMS) \cite{CopernicusLandMonitoringService} at various spatial and temporal resolutions. 

For each vegetation index, multiple data sources from the Copernicus Land Monitoring Service (CLMS) were employed, each differing in spatial and temporal resolution. Depending on the source, some datasets provide global coverage, while others are limited to the European continent. To ensure efficiency and relevance, the data were downloaded specifically for the region of interest illustrated in Figure~\ref{fig:fires_and_area_of_interest}.

In the following, a detailed description is provided of the sources selected for each index, the procedures followed during data acquisition, and the preprocessing steps applied to harmonise all variables with the target spatial resolution of 1km × 1km and the daily temporal granularity required by the \textit{IberFire} datacube. This process was done using Python, and all the vegetation indices were incorporated into the dataset as spatio-temporal features.

\subsubsection*{FAPAR}

The {Fraction of Absorbed Photosynthetically Active Radiation} (\texttt{FAPAR}) is a key biophysical variable that is commonly used as an indicator of vegetation health. It has been identified as one of the 50 Essential Climate Variables by the Global Climate Observing System (GCOS) \cite{GCOS}. In the context of wildfire risk prediction, FAPAR is particularly relevant as it reflects the photosynthetic activity and water stress level of vegetation, which are tightly linked to fuel dryness.

The CLMS provides multiple datasets for the FAPAR variable; however, no single source spans the entire temporal range required by the \textit{IberFire} datacube. To achieve full temporal coverage, two complementary datasets were integrated. The first source \cite{FAPAR_v1} offers FAPAR measurements at a spatial resolution of 1km $\times$ 1km with a 10-day temporal frequency and was used to cover the period from 01/12/2007 to 30/04/2020. The second source \cite{FAPAR_v2}, used for the remaining period from 01/05/2020 to 31/12/2024, provides enhanced spatial resolution at 300m $\times$ 300m, while maintaining the same temporal frequency.

The two datasets were harmonised as follows. For the first source, since its native resolution matched the target resolution of the datacube, values were directly aligned to the grid using nearest-neighbour resampling. In contrast, the higher-resolution values from the second source were aggregated by computing the mean FAPAR value within each corresponding 1km $\times$ 1km cell, effectively applying the average resampling method.

\subsubsection*{LAI}

The {Leaf Area Index} (\texttt{LAI}) represents the total one-sided green leaf area per unit ground area (m$^2$/m$^2$) and is a key indicator of vegetation density and structure. It is also one of the Essential Climate Variables (ECVs) identified by the GCOS, and is widely used in ecological modelling and fire risk assessment.

The download and preprocessing procedures for this variable followed the same approach as for the \texttt{FAPAR} index. A dataset with a 1 km spatial resolution was used for the period from 01/12/2007 to 30/04/2020 \cite{LAI_v1} and was aligned to the datacube using nearest-neighbour resampling. For the remaining period, from 01/05/2020 to 31/12/2024, a dataset at 300 m resolution was employed \cite{LAI_v2}, and values were aggregated to 1 km resolution using average resampling.

\subsubsection*{NDVI}
The {Normalised Difference Vegetation Index} (\texttt{NDVI}) is a remote sensing indicator that estimates vegetation health. It is calculated from the red and near-infrared spectral bands and typically ranges from $-1$ to $1$, with higher values indicating denser and healthier vegetation. In the context of wildfire risk assessment, NDVI serves as a proxy for vegetation condition and fuel availability.

As with the previous indices, two complementary datasets from the CLMS were used to ensure full temporal coverage. The first dataset \cite{NDVI_v1}, with a spatial resolution of $1 km \times 1km$ and a 10-day frequency, covers the period from 01/12/2007 to 30/06/2020. The second dataset \cite{NDVI_v2}, with a finer resolution of 300 m and the same temporal frequency, was used from 01/07/2020 to 31/12/2024. The harmonization process mirrored that used for \texttt{FAPAR} and \texttt{LAI}: nearest-neighbour resampling was applied to the 1 km dataset, while the 300 m data were aggregated to 1 km using average resampling.

\subsubsection*{LST}
Land Surface Temperature (\texttt{LST}) represents the skin temperature of the Earth's surface, expressed in Kelvin. Unlike other vegetation-related indices, \texttt{LST} is available from CLMS as an hourly variable. However, to achieve full temporal coverage for the 2007–2024 period in the \textit{IberFire} datacube, three complementary data sources were integrated, as no single dataset spans the entire time range. Specifically, the ERA5-Land skin temperature variable was used to fill in the earlier years, during which CLMS does not provide LST data.

For the period from 01/12/2007 to 10/06/2010, LST values were obtained from the aggregated ERA5-Land dataset \cite{era5_postprocessed_data}, which provides global daily average skin temperature values at a spatial resolution of 9 km $\times$ 9 km. From 11/06/2010 to 18/01/2021, a CLMS source offering hourly LST at a resolution of 5km $\times$ 5km was used \cite{LST_v2}. For each daily timestamp, the average value was calculated from retrieved hourly values. Finally, from 19/01/2021 to 31/12/2024, a second CLMS source was used \cite{LST_v3}, also with hourly values with 5km $\times$ 5km resolution, and the same daily temporal averaging method was applied.

Finally, to harmonise all three datasets with the 1km $\times$ 1km resolution of \textit{IberFire}, nearest-neighbour resampling was applied in each case, since the target resolution is finer than the original resolutions.

\subsubsection*{SWI}
The {Soil Water Index} ({SWI}) is a moisture-related indicator that estimates the percentage of water retained in the upper layers of the soil. It is derived from observations of {Surface Soil Moisture} ({SSM}) using an exponential filtering approach, giving more weight to recent measurements while smoothing the temporal signal \cite{SWI_formula}.

Equation \ref{eq:SWI_formula} calculates {SWI} for time $t_n$ based on past {SSM} measurements on times $t_i$ with $i < n$:

\begin{equation}\label{eq:SWI_formula}
 \textit{SWI}(t_n) = \frac{\displaystyle\sum_{i}^n \textit{SSM}(t_i) e^{-\frac{t_n - t_i}{T}}}{\displaystyle \sum_{i}^{n}e^ {-\frac{t_n - t_i}{T}}} \,. 
\end{equation}

The parameter $T$ regulates the influence of past observations, with smaller $T$ values assigning greater weight to recent measurements and larger $T$ values producing a more temporally smoothed index. For instance, with $T=1$, a measurement taken 10 days prior contributes with a weight proportional to $e^{\frac{-10}{1}} \approx 4.5\cdot10^{-5}$, whereas with $T=10$ the same observation has a weight proportional to $e^{\frac{-10}{-10}} \approx 0.37$

To capture moisture dynamics at different temporal scales, four {SWI} variants were downloaded and included in the datacube, corresponding to $T = 1, 5, 10, 20$ and named \texttt{SWI\_001}, \texttt{SWI\_005}, \texttt{SWI\_010}, \texttt{SWI\_020} respectively. The data were retrieved from CLMS, which provides daily {SWI} values at a spatial resolution of 12.5km $\times$ 12.5km \cite{SWI_dataset}. To align with the datacube’s 1 km $\times$ 1 km resolution, nearest-neighbour interpolation was applied during the resampling process.

\subsection{Summary of external data sources} \label{sec_2:external_data}

To ensure transparency and reproducibility, Table~\ref{tab:datasources} provides direct links to the original raw datasets used in the construction of the \textit{IberFire} datacube. Each dataset listed in the table corresponds to one or more features included in the datacube. These sources include official and publicly accessible repositories from European and national institutions, and their inclusion allows users to verify data provenance or perform additional processing tailored to specific use cases, like model deployment.

\begin{table}[htpb]
\footnotesize
\begin{tabular}{|p{1.9cm}|p{3cm}|p{1.3cm}|p{6.5cm}|}
\hline
\multicolumn{1}{|c|}{\textbf{Category}} & \multicolumn{1}{c|}{\textbf{Retrieved feature}}                                 & \multicolumn{1}{c|}{\textbf{Resolution}} & \multicolumn{1}{c|}{\textbf{Original source}}                                                                                 \\ \hline
Auxiliary features                      & Spain boundary                                                                  & Geometries                               & \scriptsize \url{https://simplemaps.com/gis/country/es} \cite{simplemaps-spain} \footnotesize                                                                                         \\ \hline
\multirow{2}{*}{Fire history}           & Historical fire data                                                            & Geometries                               & \scriptsize \url{https://forest-fire.emergency.copernicus.eu/apps/data.request.form/} \cite{effis} \footnotesize                                                           \\ \cline{2-4} 
                                        & FWI                                                                             & 27.5km                                   & \scriptsize \url{https://ewds.climate.copernicus.eu/datasets/cems-fire-historical-v1?tab=overview} \cite{FWI_data} \footnotesize                                             \\ \hline
Geographical location                   & Autonomous Communities                                                          & Geometries                               & \scriptsize \url{https://www.arcgis.com/home/item.html?id=5f689357238847bc823a2fb164544a77} \cite{area2018limites} \footnotesize                                                   \\ \hline
Land usage                              & CLC\_2006, CLC\_2012, CLC\_2018                                                 & 100m                                     & \scriptsize \url{https://land.copernicus.eu/en/products/corine-land-cover} \cite{CLC_2006_version} \cite{CLC_2012_version} \cite{CLC_2018_version}  \footnotesize                                                                      \\ \hline
Topography                              & Elevation, slope, aspect, roughness                                             & 30m                                      & \scriptsize \url{https://portal.opentopography.org/raster?opentopoID=OTSDEM.032021.4326.3} \cite{EU_DEM}   \footnotesize                                                  \\ \hline
\multirow{5}{*}{Human activity}         & Population density                                                              & 1km                                      & \scriptsize \url{https://hub.worldpop.org/doi/10.5258/SOTON/WP00674} \cite{popdens} \footnotesize                                                                              \\ \cline{2-4} 
                                        & Distance to roads                                                               & 100m                                     & \scriptsize \url{https://hub.worldpop.org/geodata/summary?id=17504} \cite{distance_to_roads} \footnotesize                                                                            \\ \cline{2-4} 
                                        & Distance to waterways                                                           & 100m                                     & \scriptsize \url{https://hub.worldpop.org/geodata/summary?id=18002} \cite{distance_to_waterways} \footnotesize                                                                             \\ \cline{2-4} 
                                        & Railways raw data                                                               & Geometries                               & \scriptsize \url{https://download.geofabrik.de/europe/spain.html} \footnotesize                                                                              \\ \cline{2-4} 
                                        & Natura 2000 network                                                             & Geometries                               & \scriptsize \url{https://www.miteco.gob.es/es/biodiversidad/servicios/banco-datos-naturaleza/informacion-disponible/rednatura\_2000\_desc.html}   \footnotesize \\ \hline
\multirow{2}{*}{Meteorological}         & 2m temperature, 2m dewpoint temperature, precipitations, 10m u-wind, 10m v-wind & 9km, hourly                              & \scriptsize \url{https://cds.climate.copernicus.eu/datasets/reanalysis-era5-land?tab=overview} \cite{era5_hourly_data} \footnotesize                                                 \\ \cline{2-4} 
                                        & 2m temperature, surface pressure, 10m u-wind, 10m v-wind                        & 9km, daily                               & \scriptsize \url{https://cds.climate.copernicus.eu/datasets/derived-era5-land-daily-statistics?tab=overview} \cite{era5_postprocessed_data}  \footnotesize                                   \\ \hline
\multirow{10}{*}{Vegetation}            & \multirow{2}{*}{FAPAR}                                                          & 1km, 10-daily                            & \scriptsize \url{https://land.copernicus.eu/en/products/vegetation/fraction-of-absorbed-photosynthetically-active-radiation-v2-0-1km} \cite{FAPAR_v1} \footnotesize          \\ \cline{3-4} 
                                        &                                                                                 & 300m, 10-daily                           & \scriptsize \url{https://land.copernicus.eu/en/products/vegetation/fraction-of-absorbed-photosynthetically-active-radiation-v1-0-300m} \cite{FAPAR_v2} \footnotesize         \\ \cline{2-4} 
                                        & \multirow{2}{*}{LAI}                                                            & 1km, 10-daily                            & \scriptsize \url{https://land.copernicus.eu/en/products/vegetation/leaf-area-index-v2-0-1km} \cite{LAI_v1} \footnotesize                                                   \\ \cline{3-4} 
                                        &                                                                                 & 300m, 10-daily                           & \scriptsize \url{https://land.copernicus.eu/en/products/vegetation/leaf-area-index-300m-v1.0} \cite{LAI_v2} \footnotesize                                                  \\ \cline{2-4} 
                                        & \multirow{2}{*}{NDVI}                                                           & 1km, 10-daily                            & \scriptsize \url{https://land.copernicus.eu/en/products/vegetation/normalised-difference-vegetation-index-v3-0-1km} \cite{NDVI_v1} \footnotesize                            \\ \cline{3-4} 
                                        &                                                                                 & 300m, 10-daily                           & \scriptsize \url{https://land.copernicus.eu/en/products/vegetation/normalised-difference-vegetation-index-v2-0-300m} \cite{NDVI_v2} \footnotesize                           \\ \cline{2-4} 
                                        & \multirow{3}{*}{LST}                                                            & 9km, daily                               & \scriptsize \url{https://cds.climate.copernicus.eu/datasets/derived-era5-land-daily-statistics?tab=overview} \cite{era5_postprocessed_data} \footnotesize                                   \\ \cline{3-4} 
                                        &                                                                                 & 5km, hourly                              & \scriptsize \url{https://land.copernicus.eu/en/products/temperature-and-reflectance/hourly-land-surface-temperature-global-v1-0-5km} \cite{LST_v2} \footnotesize           \\ \cline{3-4} 
                                        &                                                                                 & 5km, hourly                              & \scriptsize \url{https://land.copernicus.eu/en/products/temperature-and-reflectance/hourly-land-surface-temperature-global-v2-0-5km} \cite{LST_v3} \footnotesize           \\ \cline{2-4} 
                                        & SWI\_001, SWI\_005, SWI\_010, SWI\_020                                          & 12.5km, daily                            & \scriptsize \url{https://land.copernicus.eu/en/products/soil-moisture/daily-soil-water-index-global-12-5km} \cite{SWI_dataset} \footnotesize                               \\ \hline
\end{tabular}
\caption{Links to the sources of the raw data used in the construction of the \textit{IberFire} datacube.}\label{tab:datasources}
\end{table}

\newpage
\section{Data Records} \label{section:data_records}

\textit{IberFire} comprises approximately $6.8 \times 10^{9}$ individual cells, resulting from the combination of $1188$ values along the $x$ coordinate, 920 along the $y$ coordinate, and $6241$ distinct timestamps. When \textit{IberFire} is opened with the \texttt{xarray} Python package, 261 `data variables' are shown. These are all the spatio-temporal and spatial-only features that can be retrieved for each cell, including the low update frequency features stored as spatial-only features. 

Therefore, the three versions of the CLC dataset, each with 63 features, along with the 13 yearly population density features, result in a total of 261 `data variables'. However, when the correct versions of CLC and population density are selected for each cell, there are a total of $261 - 63\cdot 2 - 12 = 123$ different `real' variables. From those 123 features, the auxiliary features are not intended to be used for modelling purpose, but for data manipulation. Therefore, \textit{IberFire} provides a total of 120 different features for modelling.

Since opening the dataset initially displays the `data variables', this section presents a detailed description of them, organised into structured tables that replicate the format found on \textit{IberFire} when opened with \texttt{xarray}.

The datacube is publicly available on Zenodo (\color{blue}\url{https://zenodo.org/records/15798999}\color{black}), and the variables are organised in the following tables according to the categories defined in Section \ref{chapter:methods}. Table \ref{tab:aux_firehist_gegraph_features} presents the three auxiliary features, two fire-related variables, and five geographic context features. Table \ref{tab:clc2006} details the 63 variables derived from the 2006 CLC dataset (the 2012 and 2018 CLC version are not described since they mirror the 2006 version). Table \ref{tab:topo_human_features} then presents the 15 topographical and 21 human activity features (including the repeated population density features). Lastly, Table \ref{tab:meteo_vegetation_features} summarises the 17 meteorological variables and 8 vegetation indices included in the dataset.

\begin{table}[H]
    \centering
    \begin{tabularx}{\textwidth}{|l|X|m{3.8cm}|}
        \hline
        \textbf{Feature Name} & \textbf{Description} & \textbf{Values or Units} \\
        \hline
         \texttt{x\_index} & 
X-coordinate index values. & Integer in [0, 1187]\\
        \texttt{y\_index} & 
Y-coordinate index values. & Integer in [0, 919]\\
        \texttt{is\_spain} & Binary mask indicating the Spanish region. & 0 (outside Spain), 1 (inside Spain)
 \\
        \hline
        \texttt{is\_fire} & Binary indicator denoting whether the cell was affected by a fire on that date. & 0 (no fire), 1 (fire) \\
        \texttt{is\_near\_fire} & Binary indicator showing if the cell is within a 25×25 spatial area and a 10-day window preceding a fire event. & 0 (not near fire), 1 (near fire) \\
        \hline
        \texttt{x\_coordinate} & 
X-coordinate values in the EPSG:3035 reference system. & Metres (float)\\
        \texttt{y\_coordinate} & 
Y-coordinate values in the EPSG:3035 reference system. & Metres (float)\\
        \texttt{is\_sea} & Binary indicator denoting whether a cell lies over the open sea. & 0 (land), 1 (sea) \\
        \texttt{is\_waterbody} & Binary indicator denoting whether a cell lies over inland water (e.g., lakes, rivers).  & 0 (non-water), 1 (inland water) \\
        \texttt{AutonomusCommunity} & The Autonomous Communities code. & Label from 00 to 19 \\
        \hline
    \end{tabularx}
    \caption{Top table: auxiliary features. Middle table: fire history features. Bottom table: geographical location features.}
    \label{tab:aux_firehist_gegraph_features}
\end{table}

\begin{table}[]
\begin{tabular}{|lp{7cm}|}
\hline
\multicolumn{1}{|c|}{\textbf{Feature Name}}                             & \multicolumn{1}{c|}{\textbf{Description}}                                                                 \\ \hline
\multicolumn{2}{|c|}{\textbf{Label 3 - Raw CLC Classes (1 - 44)}}                                                                                                                   \\ \hline
\multicolumn{1}{|l|}{\texttt{CLC\_2006\_1}}                             & Proportion of the original 100m $\times$ 100m cells labelled as CLC class 1 in the 1km $\times$ 1km cell  \\ \hline
\multicolumn{1}{|c|}{(...)}                                               & \multicolumn{1}{c|}{(...)}                                                                                  \\ \hline
\multicolumn{1}{|l|}{\texttt{CLC\_2006\_44}}                            & Proportion of the original 100m $\times$ 100m cells labelled as CLC class 44 in the 1km $\times$ 1km cell \\ \hline
\multicolumn{2}{|c|}{\textbf{Label 2 - Intermediate aggregations}}                                                                                                                  \\ \hline
\multicolumn{1}{|l|}{\texttt{CLC\_2006\_urban\_fabric\_proportion}}              & Sum of \texttt{CLC\_2006\_1} - \texttt{CLC\_2006\_2}                                                                        \\ \hline
\multicolumn{1}{|l|}{\texttt{CLC\_2006\_industrial\_proportion}}                 & Sum of \texttt{CLC\_2006\_3} - \texttt{CLC\_2006\_6}                                                                        \\ \hline
\multicolumn{1}{|l|}{\texttt{CLC\_2006\_mine\_proportion}}                       & Sum of \texttt{CLC\_2006\_7} - \texttt{CLC\_2006\_9}                                                                        \\ \hline
\multicolumn{1}{|l|}{\texttt{CLC\_2006\_artificial\_vegetation\_proportion}}     & Sum of \texttt{CLC\_2006\_10} - \texttt{CLC\_2006\_11}                                                                      \\ \hline
\multicolumn{1}{|l|}{\texttt{CLC\_2006\_arable\_land\_proportion}}               & Sum of \texttt{CLC\_2006\_12} - \texttt{CLC\_2006\_14}                                                                      \\ \hline
\multicolumn{1}{|l|}{\texttt{CLC\_2006\_permanent\_crops\_proportion}}           & Sum of \texttt{CLC\_2006\_15} - \texttt{CLC\_2006\_17}                                                                      \\ \hline
\multicolumn{1}{|l|}{\texttt{CLC\_2006\_heterogeneous\_agriculture\_proportion}} & Sum of \texttt{CLC\_2006\_19} - \texttt{CLC\_2006\_22}                                                                      \\ \hline
\multicolumn{1}{|l|}{\texttt{CLC\_2006\_forest\_proportion}}                     & Sum of \texttt{CLC\_2006\_23} - \texttt{CLC\_2006\_25}                                                                      \\ \hline
\multicolumn{1}{|l|}{\texttt{CLC\_2006\_scrub\_proportion}}                      & Sum of \texttt{CLC\_2006\_26} - \texttt{CLC\_2006\_29}                                                                      \\ \hline
\multicolumn{1}{|l|}{\texttt{CLC\_2006\_open\_space\_proportion}}                & Sum of \texttt{CLC\_2006\_30} - \texttt{CLC\_2006\_34}                                                                      \\ \hline
\multicolumn{1}{|l|}{\texttt{CLC\_2006\_inland\_wetlands\_proportion}}           & Sum of \texttt{CLC\_2006\_35} - \texttt{CLC\_2006\_36}                                                                      \\ \hline
\multicolumn{1}{|l|}{\texttt{CLC\_2006\_maritime\_wetlands\_proportion}}         & Sum of \texttt{CLC\_2006\_37} - \texttt{CLC\_2006\_39}                                                                      \\ \hline
\multicolumn{1}{|l|}{\texttt{CLC\_2006\_inland\_waters\_proportion}}             & Sum of \texttt{CLC\_2006\_40} - \texttt{CLC\_2006\_41}                                                                      \\ \hline
\multicolumn{1}{|l|}{\texttt{CLC\_2006\_marine\_waters\_proportion}}             & Sum of \texttt{CLC\_2006\_42} - \texttt{CLC\_2006\_44}                                                                      \\ \hline
\multicolumn{2}{|c|}{\textbf{Label 1 - High level aggregations}}                                                                                                                    \\ \hline
\multicolumn{1}{|l|}{\texttt{CLC\_2006\_artificial\_proportion}}                 & Sum of \texttt{CLC\_2006\_1} - \texttt{CLC\_2006\_11}                                                                                   \\ \hline
\multicolumn{1}{|l|}{\texttt{CLC\_2006\_agricultural\_proportion}}               & Sum of \texttt{CLC\_2006\_12} - \texttt{CLC\_2006\_22}                                                                                  \\ \hline
\multicolumn{1}{|l|}{\texttt{CLC\_2006\_forest\_and\_semi\_natural\_proportion}} & Sum of \texttt{CLC\_2006\_23} - \texttt{CLC\_2006\_34}                                                                                  \\ \hline
\multicolumn{1}{|l|}{\texttt{CLC\_2006\_wetlands\_proportion}}                   & Sum of \texttt{CLC\_2006\_35} - \texttt{CLC\_2006\_39}                                                                                  \\ \hline
\multicolumn{1}{|l|}{\texttt{CLC\_2006\_waterbody\_proportion}}                  & Sum of \texttt{CLC\_2006\_40} - \texttt{CLC\_2006\_44}                                                                                  \\ \hline
\end{tabular}
\caption{Corine Land Cover features (corresponding to 2006). All features correspond to proportions, with values ranging from $0$ to $1$. The top part describes the 44 raw classes of the CLC dataset. The middle part corresponds to the 14 intermediate aggregation levels. The lower part of the table represents the 5 higher clustering levels of CLC. This hierarchical ordering is defined in Table~\ref{table:clc_legend}.}
    \label{tab:clc2006}
\end{table}

\begin{table}[H]
    \centering
    \begin{tabularx}{\textwidth}{|l|X|m{2.8cm}|}
        \hline
        \textbf{Feature Name} & \textbf{Description} & \textbf{Values or Units} \\
        \hline
        \texttt{elevation\_mean} & Mean elevation in the 1\,km $\times$ 1\,km grid cell. & 
Metres (float) \\
        \texttt{elevation\_stdev} & Standard deviation of elevation in the 1\,km $\times$ 1\,km grid cell. & Metres (float) \\
        \texttt{slope\_mean} & Mean slope in the 1\,km $\times$ 1\,km grid cell. & -\\
        \texttt{slope\_stdev} & Standard deviation slope in the 1\,km $\times$ 1\,km grid cell.  & -\\
        \texttt{roughness\_mean} & Mean roughness in the 1\,km $\times$ 1\,km grid cell. & -\\
        \texttt{rougness\_stdev} & Standard deviation roughness in the 1\,km $\times$ 1\,km grid cell. & -\\
        \texttt{aspect\_1} & Proportion of the aspect class 1 in the 1\,km $\times$ 1\,km grid cell. & Proportion [0,1] \\
        (...) & (...) & (...) \\
        \texttt{aspect\_8} & Proportion of the aspect class 8 in the 1\,km $\times$ 1\,km grid cell.  & Proportion [0,1] \\
        \texttt{aspect\_NODATA}  & Proportion of the aspect class NODATA in the 1\,km $\times$ 1\,km grid cell.  & Proportion [0,1] \\
        \hline
        \texttt{dist\_to\_roads\_mean} & Mean distance to roads in the 1\,km $\times$ 1\,km grid cell. & Kilometres (float) \\
        \texttt{dist\_to\_roads\_stdev} & Standard deviation of the distance to roads in the 1\,km $\times$ 1\,km grid cell. & Kilometres (float)\\
        \texttt{dist\_to\_waterways\_mean} & Mean distance to waterways in the 1\,km $\times$ 1\,km grid cell. & Kilometres (float)\\
        \texttt{dist\_to\_waterways\_stdev} & Standard deviation of the distance to waterways in the 1\,km $\times$ 1\,km grid cell. & Kilometres (float)\\
        \texttt{dist\_to\_railways\_mean} & Mean distance to railways in the 1\,km $\times$ 1\,km grid cell. & -\\
        \texttt{dist\_to\_railways\_stdev} & Standard deviation of the distance to railways in the 1\,km $\times$ 1\,km grid cell. & -\\
        \texttt{is\_holiday} & Binary mask indicating whether it is a holiday in the 1\,km $\times$ 1\,km grid cell in that time or not. & 0 (working day), 1 (holiday) \\
        \texttt{is\_natura2000} & Binary mask indicating whether the 1\,km $\times$ 1\,km grid cell is part of the Natura 2000 network or not. & 0 (is not in), 1 (is in)\\
        \texttt{popdens\_2008} & Mean population density in the 1\,km $\times$ 1\,km grid cell for the year 2008. & People/$km^2$ \\
         (...) & (...) & (...) \\
        \texttt{popdens\_2020} & Mean population density in the 1\,km $\times$ 1\,km grid cell for the year 2020. & People/$km^2$\\
        \hline
    \end{tabularx}
    \caption{Top: topography features. Bottom: human activity features.}
    \label{tab:topo_human_features}
\end{table}

\begin{table}[H]
    \centering
    \begin{tabularx}{\textwidth}{|p{4cm}|X|m{2.8cm}|}
        \hline
        \textbf{Feature Name} & \textbf{Description} & \textbf{Values or Units} \\
        \hline
        \texttt{t2m\_mean} & 
The mean temperature of air measured at 2m above the surface of the land, sea or inland waters. &   Degrees Celsius\\
        \texttt{t2m\_min} & The minimum temperature of air measured at 2m above the surface of the land, sea or inland waters. &   Degrees Celsius\\ 
        \texttt{t2m\_max} & The maximum temperature of air measured at 2m above the surface of the land, sea or inland waters. &   Degrees Celsius\\ 
        \texttt{t2m\_range} & The range temperature of air measured at 2m above the surface of the land, sea or inland waters. &   Degrees Celsius\\
        \texttt{RH\_mean} & The mean relative humidity of air. & [0, 100] in \%\\
        \texttt{RH\_min} & The minimum relative humidity of air. & [0, 100] in \%\\
        \texttt{RH\_max} & The maximum relative humidity of air. & [0, 100] in \%\\
        \texttt{RH\_range} & The range relative humidity of air. & [0, 100] in \%\\
        \texttt{surface\_pressure\_mean} & The mean surface pressure of air. & Hectopascal \\
        \texttt{surface\_pressure\_min} & The minimum surface pressure of air. & Hectopascal \\
        \texttt{surface\_pressure\_max} & The maximum surface pressure of air. & Hectopascal \\
        \texttt{surface\_pressure\_range} & The range surface pressure of air. & Hectopascal \\
        \texttt{total\_precipitation} \texttt{\_mean} & Mean of the hourly values of the total precipitation variable from ERA5-Land. & Millimetres (l/$m^2$)\\
        \texttt{wind\_speed\_mean} & 
The mean wind speed derived from the hourly u-component and v-component of wind. & Metres per second (m/s)\\
        \texttt{wind\_speed\_max} & The maximum wind speed.  & Metres per second (m/s)\\
        \texttt{wind\_direction\_mean} & The mean wind direction (where the wind comes). & Degrees\\
        \texttt{wind\_direction\_at} \texttt{\_max\_speed} & The wind direction (where the wind comes) where the maximum wind speed happened. & Degrees\\
        \hline
        \texttt{FAPAR} & 
Fraction of Absorbed Photosynthetically Active Radiation, the fraction of the solar radiation absorbed by live plants for photosynthesis. & -\\
        \texttt{LAI} & Leaf Area Index, representing the half of the total green canopy area per unit horizontal ground area. & -\\
        \texttt{LST} & 
Land Surface Temperature is the temperature of the surface of the Earth. & Degrees Kelvin \\
        \texttt{ndvi} &  
Normalised Difference Vegetation Index, an indicator of the greenness of the biomes. & - \\
        \texttt{SWI\_001} & Soil Water Index at T=1, the moisture humidity conditions of the soil. & [0, 100] in \%\\
        \texttt{SWI\_005} & Soil Water Index at T=5, the moisture humidity conditions of the soil.  & [0, 100] in \% \\
        \texttt{SWI\_010} & Soil Water Index at T=10, the moisture humidity conditions of the soil.  & [0, 100] in \%\\
        \texttt{SWI\_020} & Soil Water Index at T=20, the moisture humidity conditions of the soil.  & [0, 100] in \%\\       
        \hline
    \end{tabularx}
    \caption{Top: meteorological features. Bottom: vegetation indices.}
    \label{tab:meteo_vegetation_features}
\end{table}

\newpage

\section{Technical Validation} \label{sec:technical_validation}

Four distinct validation procedures were conducted to ensure the robustness and reliability of \textit{IberFire}. First, the integrity of data formats and measurement units was verified to guarantee internal consistency across all features. Second, given that vegetation indices were derived from satellite-based remote sensing data, the presence of missing values was common. To address this, multiple imputation techniques were evaluated, and the method yielding the lowest reconstruction error was applied to fill missing data. Third, available AEMET historical measurement data were downloaded to compare with \textit{IberFire} meteorological data. Finally, an XGBoost model was trained on data from 2008 to 2023 and using the predicted class probabilities, fire risk maps were plotted for various 2024 days, which served as a practical validation step. This section provides a detailed account of each of these four validation strategies aimed at guaranteeing the quality and usability of the \textit{IberFire} datacube.

\subsection{Data correctness}

A datacube is a multi-dimensional data structure that integrates spatial and temporal information in a unified framework. Within the \textit{IberFire} datacube, this includes grid coordinate values, binary and categorical attributes, proportion-based data, and continuous numerical features. Given the complexity and heterogeneity of these components, quality assurance procedures are needed to ensure the reliability and internal consistency of the datacube. 

First, a temporal consistency check was conducted to guarantee that there are no duplicated time coordinates in the datacube. The dataset was verified to contain valid accessible values for all spatio-temporal features across the entire temporal range from 01/12/2007 to 31/12/2024. For the spatial features, visual inspection was performed to ensure that spatial coordinates are homogeneous across the entire region of interest.

For categorical and binary features, it was ensured that all entries conformed to the expected set of values. For the \texttt{AutonomousCommunities} feature, all labels were verified to fall within the valid integer range of 0 to 19, with no occurrences of invalid or unexpected values. Regarding binary features, \texttt{is\_spain}, \texttt{is\_sea}, \texttt{is\_waterbody}, \texttt{is\_holiday}, \texttt{is\_natura2000}, \texttt{is\_fire}, and \texttt{is\_near\_fire}, it was confirmed that all values were strictly limited to either 0 or 1.

For features representing proportions, such as the nine derived from aspect and the 189 calculated from CLC, data validation procedures ensured that all values lie within the range $[0,1]$, with no values falling outside this interval. Similarly, features derived from relative humidity and the four indices based on the Soil Water Index (SWI), which represent percentages, were verified to exclusively contain numerical values within the interval $[0,100]$.

Additionally, for the proportion-based features derived from the CLC dataset, structured hierarchically into three levels described in Table \ref{table:clc_legend}, consistency checks were performed to ensure that the proportions at each hierarchical level sum to one for every cell. Therefore, it was verified that the five aggregated features at Label 1, the fourteen intermediate-level features at Label 2, and the forty-four fine-grained features at Label 3 each sum to exactly one per cell.

Finally, for numerical features such as meteorological variables and vegetation indices, visual inspection was carried out through exploratory plotting to identify potential outliers and ensure that all values fell within logical and expected ranges. This process revealed a substantial number of missing values in certain data sources used to construct the vegetation indices. For instance, in the case of \texttt{FAPAR}, no missing values were observed until 30/04/2020, which corresponds to the ending of the first data source \cite{FAPAR_v1}. After that date, the second source was used \cite{FAPAR_v2} and exhibited a considerable number of gaps. The methodology adopted to address this issue is detailed in the following subsection.

\subsection{Missing values validation}
Once all features were validated to fall within their expected formats and value ranges, the missing values identified in the vegetation indices were addressed. This step was crucial, as not all machine learning algorithms can inherently manage missing data.

The imputation process was applied exclusively to the vegetation indices, since all other data sources were either complete or contained a negligible amount of missing values. In particular, no missing data were observed in any of the spatial-only features, and regarding spatio-temporal features, the layers \texttt{is\_fire}, \texttt{is\_near\_fire}, and \texttt{is\_holiday} were complete for the entire temporal range. Likewise, the meteorological variables derived from the ERA5-Land dataset did not exhibit missing values within the Spanish territory (\texttt{is\_spain} = 1), due to the spatial continuity and post-processing of the source data.

Four interpolation techniques were analysed to address the missing values in the vegetation indices: nearest neighbour, linear, quadratic, and cubic. Each method was applied along the temporal axis independently on each spatial cell.

\begin{figure}[!h]
    \centering
    \includegraphics[width=\linewidth]{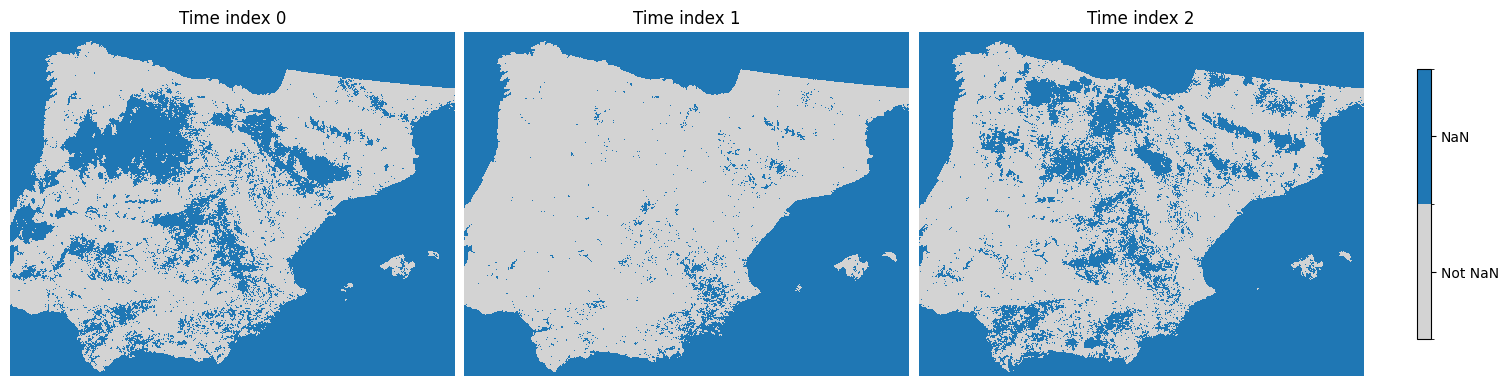}
    \vspace{0.5cm} 
    \includegraphics[width=\linewidth]{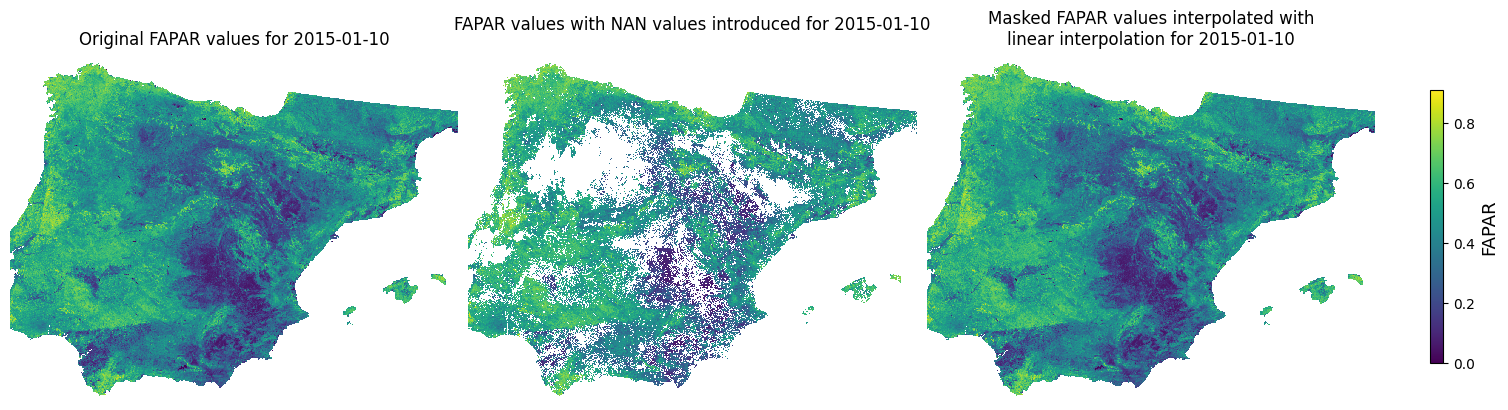} 
    \caption{Top: Three examples of binary NAN masks that were introduced as artificial NAN values. Bottom: Comparison between original FAPAR values, masked FAPAR values and interpolated FAPAR values.}
    \label{fig:nan_gap_examples}
\end{figure}

To evaluate these methods, artificial gaps were first added in dates known to be complete. These gaps followed the real shapes of the missing values observed in satellite-derived vegetation indices. Specifically, 140 NAN masks were obtained from the FAPAR variable and inserted on all the indices. Three examples of these artificial masks are shown in the top image of Figure \ref{fig:nan_gap_examples}.

Subsequently, the four imputation methods were applied to reconstruct the artificially masked values. To evaluate the reconstruction accuracy, the absolute differences between the true and the imputed values were computed for each masked day. These differences were then summed across all data points for that day, and the resulting sums were averaged over all masked days (i.e., they were divided by 140). This yielded the \textit{Reconstruction Error} (RE), which served as the primary metric for comparing the performance of the imputation methods.

The evaluation was performed independently for each vegetation index.  To illustrate the effect of the interpolation, the bottom image in Figure \ref{fig:nan_gap_examples} displays an example with the original data of FAPAR, the artificially masked version, and the reconstruction obtained using linear interpolation.

Table \ref{tab:reconstruction_mae} presents the RE scores for each vegetation index across the four imputation methods. As shown in the table, linear interpolation yielded the lowest RE across every tested feature. Consequently, it was adopted as the imputation method for filling the real missing values in the vegetation indices. It should be noted that the RE varies a lot across different vegetation indices due to the difference in magnitudes, not because some indices have inherently more missing data or are harder to reconstruct. For example the NDVI ranges between -1 and 1 whereas the SWI ranges from 0 to 100. Furthermore, RE values are not expected to match the magnitude as the original feature values, since the RE represents the sum of absolute errors across all cells for a given day.

\begin{table}[ht]
\centering
\begin{tabular}{|c|c|c|c|c|}
\hline
\textbf{Tested variable} & \textbf{Nearest neighbour} & \textbf{Linear} & \textbf{Quadratic} & \textbf{Cubic} \\
\hline
FAPAR & 3412 & \textbf{2418} & 4089 & 4191 \\
\hline
LAI & 14705 & \textbf{9941} & 14716 & 15037 \\
\hline
NDVI & 4239 & \textbf{3588} & * & * \\
\hline
SWI\_001 & 724876 & \textbf{649175} & 1507907 & 1601710 \\
\hline
SWI\_005 & 326488 & \textbf{273457} & 585050 & 608277 \\
\hline
SWI\_010 & 224292 & \textbf{177245} & 367986 & 381461 \\
\hline
SWI\_020 & 151797 & \textbf{112792} & 214688 & 222810 \\
\hline
LST & 325655 & \textbf{285057} & * & * \\
\hline
\end{tabular}
\caption{Reconstruction Error (RE) of the interpolation techniques for all the tested variables \newline(* The interpolation technique was not assessed for that feature due to execution errors caused from characteristics of the data).}
\label{tab:reconstruction_mae}
\end{table}

\subsection{Data comparison with AEMET}
The \textit{IberFire} datacube was constructed to support daily-scale applications that use meteorological data from AEMET. Given that the meteorological variables were derived from ERA5-Land data and subsequently transformed to align with the format of AEMET observations, a validation process was carried out to evaluate the accuracy of these transformations.

This validation compared the transformed ERA5-Land variables included in \textit{IberFire} against historical records from AEMET meteorological stations. Particular attention was paid to the \textit{u-wind} and \textit{v-wind} components, which underwent the most complex transformations, as they required the reconstruction of wind speed and direction.

To perform the comparison, historical AEMET data were retrieved from \color{blue}\url{https://datosclima.es/Aemethistorico/Descargahistorico.html}\color{black}. This source provides daily records from meteorological stations, including maximum, mean, and minimum temperature; precipitation; mean wind speed; wind direction at maximum wind speed; maximum wind speed; and both maximum and minimum surface pressure. Unfortunately, this source does not include historical records for other meteorological variables featured in \textit{IberFire}, such as relative humidity. No alternative source was found that offers historical records for these variables. Furthermore, the retrieved AEMET dataset contained missing values for the available variables, and many of those features were not provided in all stations.

\begin{figure}
    \centering
    \includegraphics[width=\linewidth]{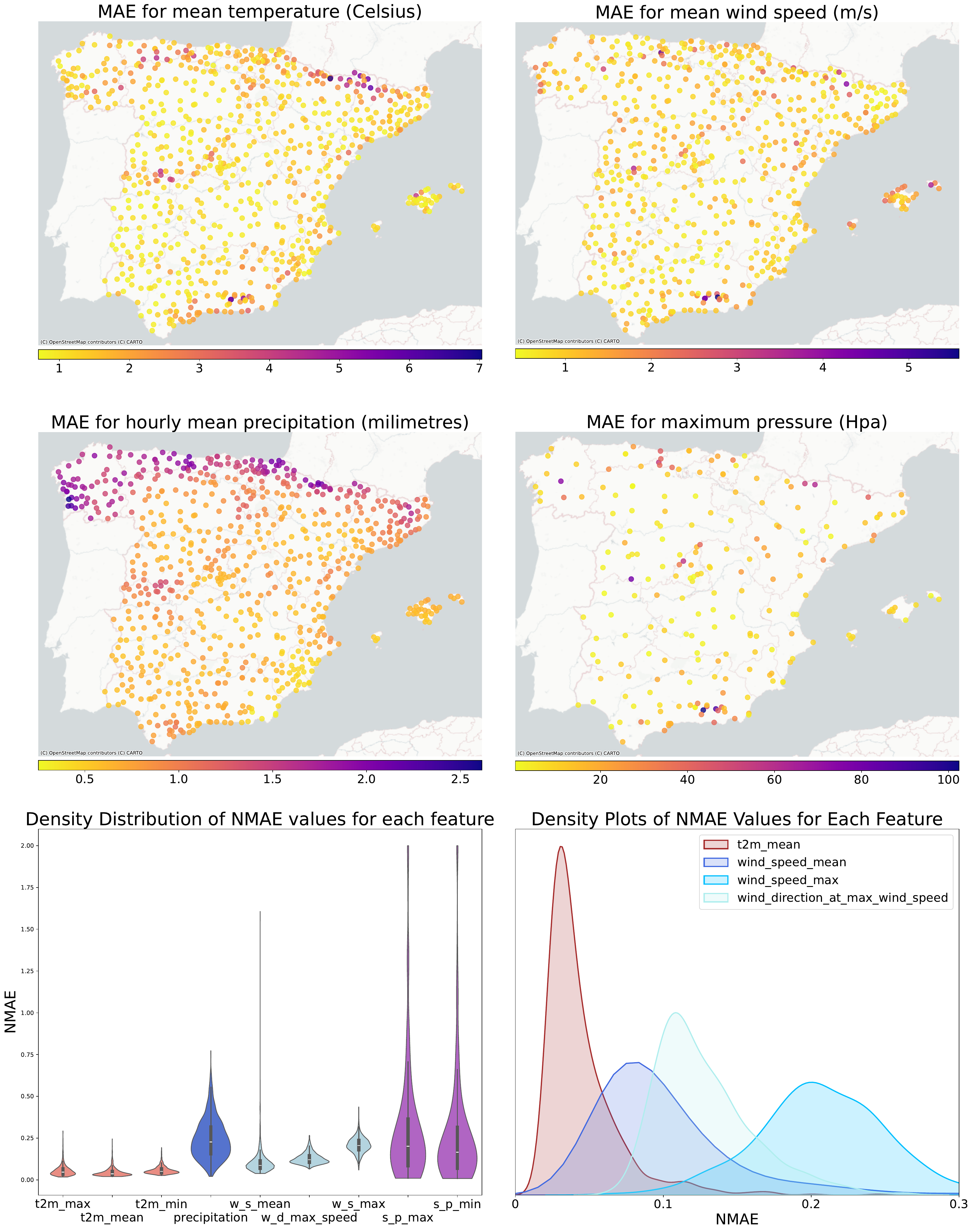}
    \caption{Four upper images: MAE values in the available stations for \texttt{t2m\_mean}, \texttt{wind\_speed\_mean}, \texttt{total\_precipitation\_mean}, \texttt{surface\_pressure\_max}. Bottom left: violin plot with NMAE values for all variables. Bottom right: density plot with NMAE values for the features with the most distinct NMAE distributions.}
    \label{fig:AEMET_validation_results}
\end{figure}

For each meteorological station within the region of interest, the Mean Absolute Error (MAE) was calculated for all available dates between 01/01/2007 and 31/12/2024, for each available feature. To calculate it, the values of the nearest cell to each meteorological station were selected. Chosen examples of these results are shown in the top four images of Figure~\ref{fig:AEMET_validation_results}. As illustrated, precipitation records are available for a considerably larger number of stations compared to surface pressure. Additionally, some stations have higher MAE values than others for the same feature.

Given the substantial regional variability in meteorological conditions, for example, the significantly higher precipitation levels in northern Spain compared to the south, a normalisation step was introduced to enable consistent comparisons of error magnitudes across features and stations. To this end, the Normalised Mean Absolute Error (NMAE) was computed according to the following expression:

\begin{equation}
    \text{NMAE(feature, station)} = \frac{\text{MAE(feature, station)}}{max(\text{feature, station}) - min(\text{feature, station})}\,,
\end{equation}
where $\text{MAE(feature, station)}$ is the MAE value of a feature in a given station, $max(\text{feature, station})$ is the maximum value of the feature measured in that station, and $min(\text{feature, station})$ is the minimum. Table \ref{tab:AEMET_validation_summary_table} provides the mean MAE and NMAE values across all stations for every available feature, and the bottom two images of Figure \ref{fig:AEMET_validation_results} visually compare the NMAE values with violin and density plots.

\begin{table}[htbp]
    \centering
    \scriptsize
    \begin{tabular}{lccccccccc}
        \toprule
         & t2m\_max & t2m\_mean & t2m\_min & precip. & w\_s\_mean & w\_d\_max\_s & w\_s\_max & s\_p\_max & s\_p\_min \\
        \midrule
        mean\_MAE   & 2.173 & 1.495 & 1.829 & 0.918 & 1.046 & 46.115 & 5.856 & 13.995 & 13.738 \\
        stdev\_MAE  & 1.215 & 0.815 & 0.732 & 0.422 & 0.563 & 11.841 & 1.728  & 15.546 & 15.531 \\
        mean\_NMAE  & 0.055 & 0.045 & 0.058 & 0.246 & 0.104 & 0.128 & 0.209    & 0.356    & 0.312 \\
        stdev\_NMAE & 0.032 & 0.026 & 0.024 & 0.118 & 0.08  & 0.033 & 0.047    & 0.629    & 0.603 \\
        \bottomrule
    \end{tabular}
    \caption{Mean and standard deviation of MAE and NMAE values across all stations. Feature names from left to right: \texttt{t2m\_max}, \texttt{t2m\_mean}, \texttt{t2m\_min}, \texttt{total\_precipitation\_mean}, \texttt{wind\_speed\_mean}, \texttt{wind\_direction\_at\_max\_speed}, \texttt{wind\_speed\_max}, \texttt{surface\_pressure\_max}, and \texttt{surface\_pressure\_min}.}
    \label{tab:AEMET_validation_summary_table}
\end{table}

The results indicate that the temperature features are highly reliable. Even in the worst-performing station, the mean Mean Absolute Error (MAE) remains as low as $2.173$, with a standard deviation of $1.215$. Assuming a normal distribution of errors, this implies that in 95\% of the cases, the discrepancy between AEMET measurements and \textit{IberFire} data does not exceed 4\degree. Given that temperature values at a single station can vary by up to 35\degree during the year, this error value is relatively small and supports the robustness of the temperature variables in the \textit{IberFire} datacube. Moreover, as the {temperature} values are derived from ERA5-Land, a dataset rigorously validated \cite{ERA5_validation} by the Copernicus Climate Data Store, their reliability is further reinforced.

In comparison, the {precipitation} and {surface pressure} features exhibit higher NMAE values than {temperature}. Nevertheless, these variables are also sourced from Copernicus ERA5-Land without suffering much transformations. Consequently, despite their relatively higher error metrics, the integrity and reliability of these features remain supported by the quality and consistency of the underlying data source.

Finally, the {wind}-related features yield intermediate NMAE values. These variables underwent the most substantial transformations, involving the reconstruction of magnitudes and angles from vector components. Despite the complexity of the transformations, the resulting error values remain within acceptable bounds, suggesting that the transformation procedures were effective and that the wind features in the \textit{IberFire} datacube are suitable for modelling applications.

\subsection{Fire risk mapping validation}\label{sec:fire_risk_mapping}

To evaluate the practical applicability of the \textit{IberFire} datacube in real-world scenarios, a fire risk mapping exercise was performed. The objective was to assess whether the datacube features can support the generation of reliable fire risk maps when used in combination with standard machine learning techniques.

Forest fires are inherently rare events, resulting in a highly imbalanced distribution of the \texttt{is\_fire} feature, with significantly fewer fire instances relative to non-fire occurrences. To address this imbalance during model training, a balanced dataset was retrieved from \textit{IberFire} by including all fire instances recorded between 2008 and 2023, complemented by an approximately equal number of randomly selected non-fire instances from the same dates. Due to the stochastic nature of the sampling process of non-fire instances, the final class counts were not exactly equal. The resulting training dataset consisted of 140,399 instances, comprising 70,476 fire cases and 69,923 non-fire cases. For each selected instance corresponding to a unique spatio-temporal cell, all input features from the previous day (selecting the adequate CLC and \texttt{popdens} versions) were extracted from the \textit{IberFire} datacube and stored in CSV format, and as target variable the \texttt{is\_fire} of the selected instance was retrieved. This one-day difference between the input features and the output variable ensures that predictions can be made using information from the previous day.

Following construction of the dataset, an XGBoost classifier was trained on the extracted instances. To reduce the risk of overfitting and ensure the generalisation capability of the model, a cross-validation strategy was applied. After model development, an independent test set was generated by retrieving a new balanced dataset, consisting of all available fire instances from the year 2024 along with a comparable number of randomly sampled non-fire instances from the same dates. The trained model was then evaluated on this test set, yielding an accuracy of 86\%. Furthermore, the model was also tested on all the instances of 2024 (around $3.1 \times 10^9$ instances) and achieved an Area Under the Receiver Operating Characteristic (AUROC) of 0.95, which indicates a high degree of predictive accuracy and robustness in real-world forecasting scenarios.

\begin{figure}[htbp]
    \begin{minipage}[b]{0.49\linewidth}
        \raggedright
        \includegraphics[width=\linewidth]{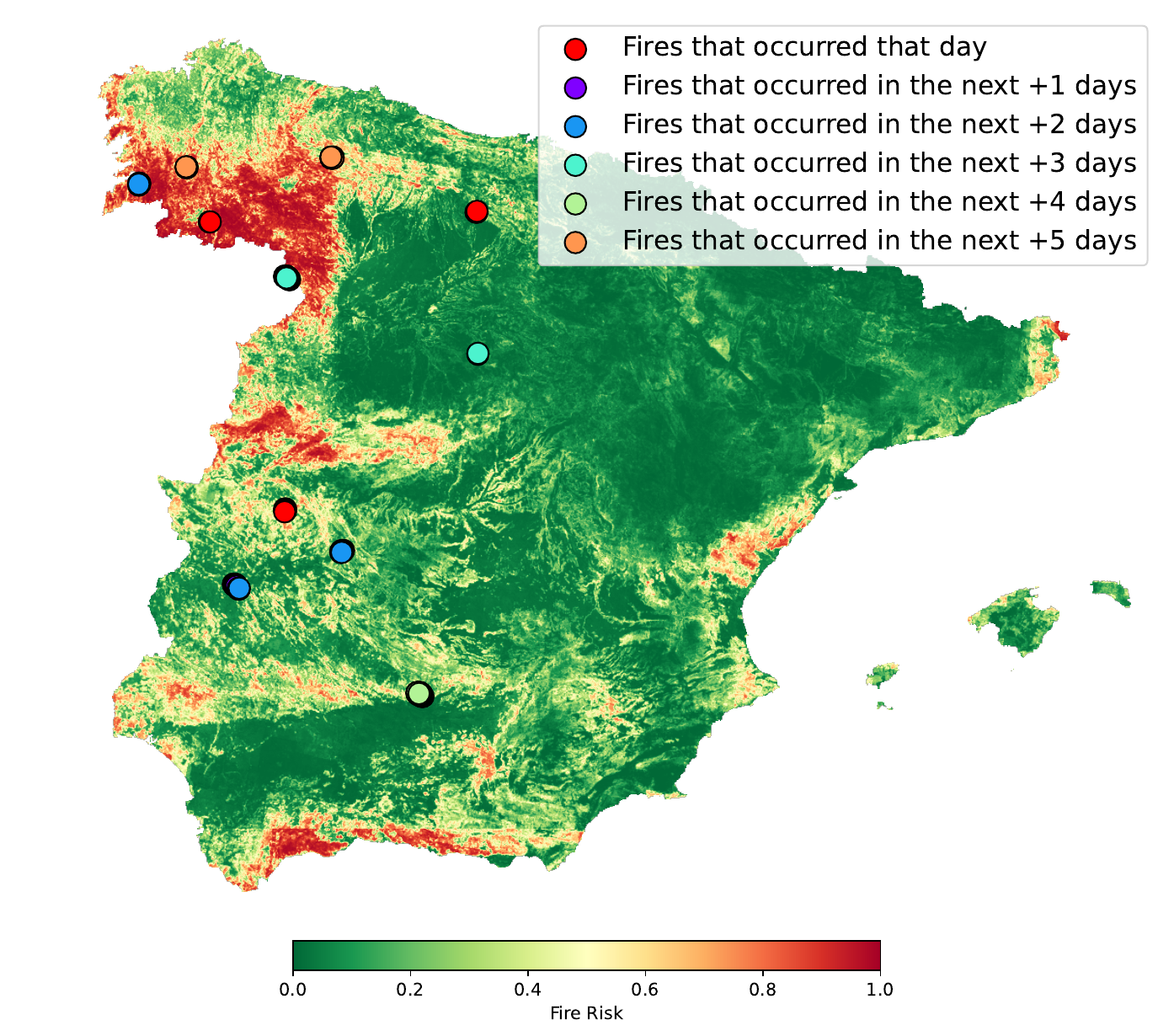}
    \end{minipage}
    \hfill
    \begin{minipage}[b]{0.49\linewidth}
        \raggedleft
        \includegraphics[width=\linewidth]{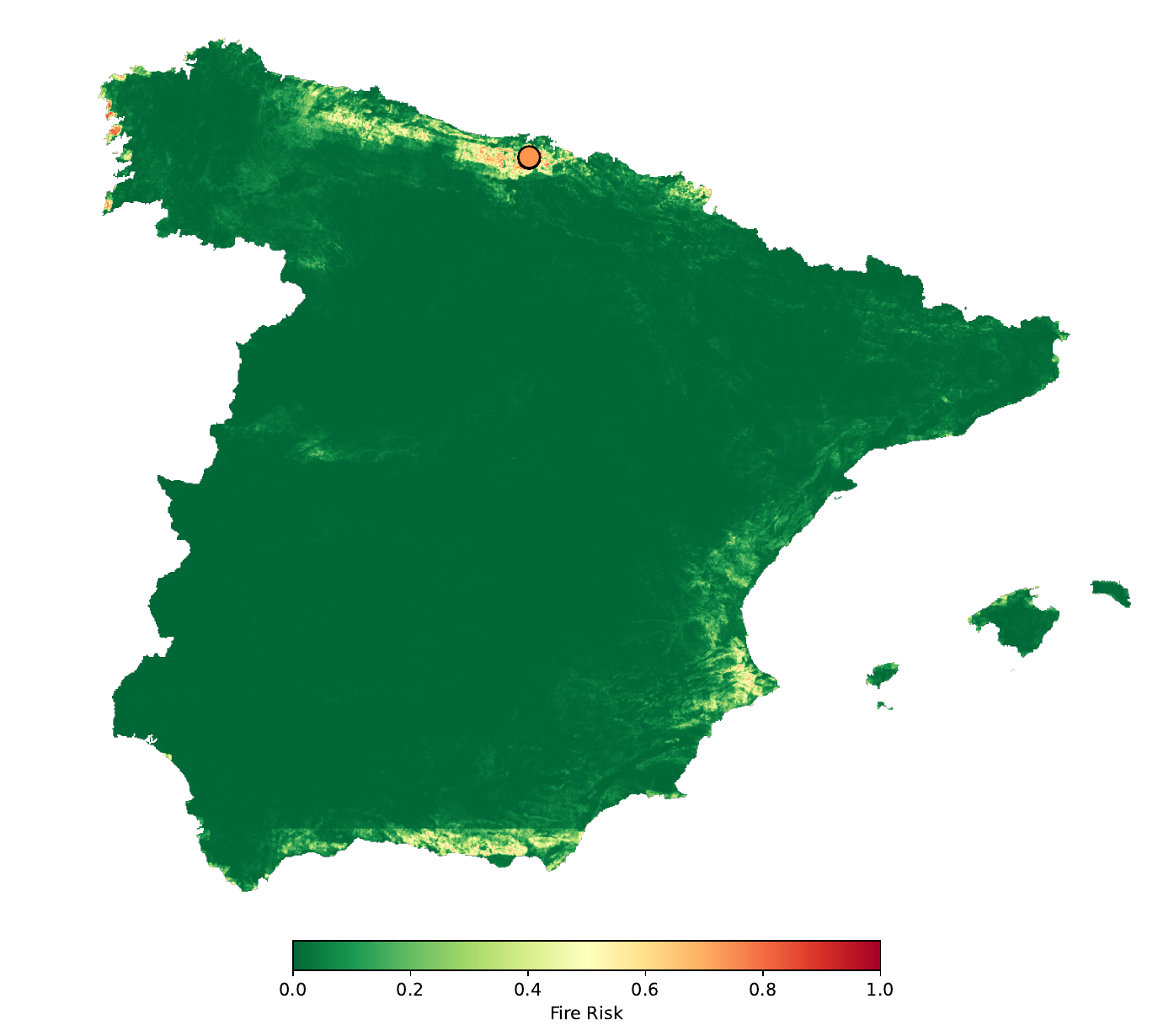}
    \end{minipage}
    \caption{Example of fire risk prediction maps. Left: predictions for 13/07/2024. Right: predictions for 20/12/2024.}
    \label{fig:firerisk_maps}
\end{figure}

In addition to this quantitative evaluation, the classifier was used to generate daily fire risk predictions for the entire year of 2024. For each day, all corresponding spatio-temporal instances were extracted from the \textit{IberFire} datacube and converted into CSV format to serve as input for the model. Predictions were then computed for each instance. Subsequently, the auxiliary features \texttt{x\_index} and \texttt{y\_index} were used to construct the daily fire risk raster maps. Figure~\ref{fig:firerisk_maps} presents a selection of predicted fire risk maps overlaid with the actual forest fire occurrences, enabling visual comparison between the predicted risk predictions and observed fires. Furthermore, the public repository associated with \textit{IberFire} presents all the fire risk maps as an animation (see Section \ref{chapter:code} for code availability).

Visual inspection of the resulting fire risk maps revealed a high degree of spatial and temporal coherence and alignment with historical fire incidence patterns illustrated in Figure~\ref{fig:fires_and_area_of_interest}. Notably, areas with historically high fire activity, such as Galicia, Asturias, and regions along the Portuguese border, consistently exhibited higher predicted risk levels, reinforcing the model’s ability to capture meaningful spatial trends in fire susceptibility. Furthermore, as Figure \ref{fig:firerisk_maps} shows, the model is capable of making accurate predictions and assigning high risk to areas where forest fires actually occur. This is particularly evident in the winter predictions, which demonstrate the model’s ability to assign generally low risk across most regions while accurately identifying localised areas with a high likelihood of fire ignition.

These results suggest that the \textit{IberFire} datacube serves as a reliable foundation for downstream predictive modelling and geospatial analysis. Fire risk mapping using machine learning models trained on \textit{IberFire} can therefore be considered a promising approach for anticipating wildfire-prone areas.

\section{Usage notes} \label{chapter:usage_notes}

The \textit{IberFire} datacube was developed to support the modelling of wildfire occurrence risk across the Spanish territory, excluding the Canary Islands, Ceuta, and Melilla. A spatio-temporal structure was implemented to fulfil this objective, providing daily data from 01/12/2007 to 31/12/2024 over a regular grid of 1km $\times$ 1km spatial resolution. This design enables the training and validation of ML and DL models for wildfire risk prediction. The datacube is intended exclusively for mainland Spain and the Balearic Islands. To ensure that only relevant cells with complete and valid data are included in modelling workflows, users should filter the dataset accordingly using the variable \texttt{is\_spain} (by selecting the instances where \texttt{is\_spain} = 1).

The datacube includes daily records beginning in December 2007; however, fire occurrence data were retrieved only from 01/01/2008 onward, and all cells for December 2007 were explicitly set as non-fire. This initial month is included to support the modelling of DL models that require a sequence of previous dates, like long short-term memory (LSTM) networks. Thus, even though fire predictions focus on the period from 2008 to 2024, the extended time frame ensures that models requiring temporal dependencies can be effectively trained.

Prediction targets can be defined using either the \texttt{is\_fire} variable, which indicates whether a fire occurred in a given cell on a specific date, or the \texttt{is\_near\_fire} variable, which flags cells located in proximity to a fire event. The use of the auxiliary variables \texttt{is\_spain}, \texttt{x\_index} and \texttt{y\_index} as model inputs is not recommended. These features were introduced to facilitate data filtering, but they lack geospatial meaning beyond the cell grid. Their values are specific to the internal indexing of the datacube and are not transferable to other geographic regions. For modelling spatial geographical location, instead of \texttt{x\_index} and \texttt{y\_index}, the variables \texttt{x\_coordinate} and \texttt{y\_coordinate}, which represent the actual projected coordinates of each grid cell on the EPSG:3035 CRS, can be used as they retain geographic meaning and are suitable for location-aware modelling. 

The remaining features in the datacube are suitable for use as explanatory variables in predictive models. Users can decide whether to integrate the baseline model \texttt{FWI} as an input feature, which when included, would result in a total of 120 different features for fire risk modelling.

To incorporate information about the Spanish territorial division, it is recommended to apply one-hot encoding to the categorical \texttt{AutonomousCommunities} feature. This results in 17 binary features corresponding to the autonomous communities listed in Table \ref{tab:community_codes}, excluding the Canary Islands, Ceuta, and Melilla, which are not part of the study area. Similarly, a \texttt{month} feature can be derived from the temporal coordinate of each spatio-temporal cell. As different months exhibit distinct seasonal fire risk patterns, it is also advisable to apply one-hot encoding to this variable.

It is important to note that some features, such as those derived from CLC and population density (\texttt{popdens}), are not static but also not temporally continuous. These variables are associated with a specific year of update and thus are included in the datacube with spatial dimensions only $(x,y)$, rather than full spatio-temporal indices $(x, y,t)$. When preprocessing the datacube for modelling applications or conversion to tabular formats such as CSV or DataFrame structures, it is crucial to ensure that for each spatio-temporal cell $(x,y,t)$, only information available at the time $t$ of the cell is used. For instance, when modelling a record for July 15, 2010, the appropriate population density value to use is \texttt{popdens\_2010}, which should be assigned to a generic \texttt{popdens} column. Similarly, for a CLC-derived feature such as \texttt{CLC\_urban\_fabric\_proportion}, the correct value to assign would be taken from \texttt{CLC\_2006\_urban\_fabric\_proportion}, since the next update in 2012 would not yet be available at that date. Careful consideration is required to prevent potential data leakage, particularly in the case of land cover features that might implicitly reflect fire occurrence. The CLC dataset contains a class representing the area that was burned, which, if taken from a dataset version updated after the date being modelled, could inadvertently introduce information correlated with the target variable (\texttt{is\_fire}). This leakage could compromise the integrity and validity of predictive modelling results.

To create practically useful models, the instance extraction process should leverage historical data. For instance, the feature values from one day can be used to predict the value of the \texttt{is\_fire} feature for the following day. For more sophisticated models, such as LSTM networks, a larger window of historical data can be used.

For real-world model deployment, relying on ERA5-Land as the meteorological data source is not feasible due to its inherent 5-day latency. To address this limitation, the construction of the \textit{IberFire} datacube was carefully designed taking into consideration operational applicability. Consequently, all meteorological features were selected and processed to align with the format and units of the open-access, near-real-time data provided by AEMET. This compatibility was validated in Section \ref{sec:technical_validation} through a comparison between historical AEMET measurements and the corresponding \textit{IberFire} records. While alternative meteorological sources could be employed for model deployment instead of AEMET, doing so would require careful preprocessing to ensure compatibility with the trained models.

Class imbalance presents a significant challenge in training models for forest fire risk prediction. The proportion of positive fire instances is extremely low, as the likelihood of a specific area burning on a given day is, in general, minimal. To mitigate this imbalance during the training phase, it is advisable to construct a balanced dataset by incorporating all fire occurrences and randomly sampling an equal number of non-fire (and non near-fire) instances.

Additionally, Figure \ref{fig:fires_and_area_of_interest} displays all recorded fires during the \textit{IberFire} study period, overlaid across the territory. As observed, some autonomous communities exhibit a significantly higher number of fire records than others. This spatial disparity may be attributed to various natural factors, such as regional climatic conditions, or anthropogenic factors, including local legislation or land management practices. Regardless of the cause, it is essential to account for this variability when designing predictive models. In certain cases, it may be advisable to develop separate models for regions sharing similar environmental or regulatory characteristics.

Directly comparing the trained models with the baseline Fire Weather Index (FWI) is not straightforward, as AI classifiers typically produce probabilistic outputs, while the FWI generates continuous, regression-like values. To enable classification using the FWI, it is necessary to discretize its outputs. Based on the thresholds in Table \ref{tab:FWI_risk_levels}, FWI values can be clipped at a maximum of 50 and linearly scaled to the [0, 1] range to approximate probabilities. Using this approach, cells with FWI values below 25 are classified as ‘non-fire’, while those above 25 are classified as ‘fire’, in accordance with the fire risk categories.

An essential final consideration when working with the \textit{IberFire} datacube is the substantial computational resources required for data processing. Although the total disk size of the datacube is approximately 29GB, each spatio-temporal feature is stored in a highly compressed format. When decompressed, a single feature represented as a \texttt{float32} array occupies around 25GB of memory, making it impractical to load multiple features into RAM simultaneously on standard hardware. The \texttt{xarray} package handles this problem by chunking the data. For lightweight tasks such as data visualization or exploratory analysis, standard hardware is sufficient. Similarly, retrieving a balanced CSV training dataset, including all fire instances of \textit{IberFire} and an equal number of non-fire instances, can be done with standard hardware, as well as generating daily fire risk maps with all the daily instances. However, more demanding operations, such as the addition or generation of new spatio-temporal variables derived from combinations of existing features, require significantly more memory. Therefore, all processing steps involved in the construction of \textit{IberFire}, including preprocessing and training dataset extraction, were executed on a machine equipped with 128GB of RAM.

\section{Code Availability}\label{chapter:code}

The functions for loading and transforming the data were implemented in Python. The processing code used to generate the \textit{IberFire} datacube, to validate it and to visualize it, is available on GitHub \color{blue}\url{https://github.com/JulenErcibengoaTekniker/IberFire}\color{black}.

\section*{Author Information}

\subsection*{Authors and Affiliations}
\textbf{Intelligent Information Systems Unit, Tekniker, 20600, Eibar, Spain}\\
\indent Julen Ercibengoa, Meritxell Gómez-Omella
\\~\\
\textbf{Department of Languages and Computer Systems, University of the Basque Country (UPV/EHU), Donostia-San Sebastián, Spain}\\
\indent Julen Ercibengoa, Izaro Goienetxea

\subsection*{Contributions}
\textbf{Julen Ercibengoa:} Conceptualisation, Methodology, Software, Data source finding, Data retrieval, Data curation, Visualisation, Writing – original draft. 
\\~\\
\textbf{Meritxell Gómez-Omella:} Conceptualisation, Project administration, Supervision Methodology, Supervision, Writing – review \& editing. 
\\~\\
\textbf{Izaro Goienetxea:} Conceptualisation, Supervision Methodology, Supervision, Writing – review \& editing.

\subsection*{Corresponding Author}

Correspondace to Julen Ercibengoa\\
julen.ercibengoa@tekniker.es

\section*{Competing Interests}
The authors declare no competing interests.

\section*{Acknowledgments}

This work was partly supported by GAIA project (Ref PLEC2023-010303) “Gestión integral para la prevención, extinción y reforestación debido a incendios forestales” from Spanish Research Agency (AEI). 

\newpage

\bibliographystyle{naturemag}
\bibliography{bibliography_corrected}

\end{document}